%% file: root.tex
\documentclass[lettersize,journal]{IEEEtran}

\usepackage{amsmath,amsfonts}
\usepackage{algorithmic}
\usepackage{algorithm}
\usepackage{array}
\usepackage{textcomp}
\usepackage{stfloats}
\usepackage{url}
\usepackage{verbatim}
\usepackage{graphicx}
\usepackage{enumitem}
\usepackage{cite}
\usepackage{multicol,multirow}
\usepackage{booktabs}       %
\usepackage{amsfonts}       %
\usepackage{nicefrac}       %
\usepackage{microtype}      %
\usepackage[dvipsnames]{xcolor}         %
\usepackage{colortbl}
\usepackage{titletoc}
\usepackage{color}
\usepackage{caption, subcaption, overpic, textpos}
\usepackage{titlesec}
\usepackage{url}
\usepackage{setspace}
\usepackage{makecell}
\usepackage{tasks}
\usepackage{pifont}
\usepackage{framed}

\usepackage{xspace}

\usepackage{tcolorbox}

\definecolor{baselinecolor}{gray}{.9}
\definecolor{yellow}{RGB}{218,165,32}
\definecolor{LightSlateBlue}{RGB}{70,130,180}
\definecolor{DeepBlue}{RGB}{108,135,176}
\definecolor{DeepPurple}{RGB}{136,105,160}
\definecolor{LightGreen}{RGB}{117, 175, 158}
\definecolor{LightRed}{RGB}{227,120,117}

\definecolor{robo-orange}{RGB}{221, 160, 98}
\definecolor{deepblue}{RGB}{33, 95, 154}
\definecolor{cvprblue}{rgb}{0.21,0.49,0.74}
\usepackage[pagebackref,breaklinks,colorlinks,citecolor=cvprblue]{hyperref}
\usepackage{cleveref}

\newcommand{\change}[1]{\textcolor{black}{#1}}
\newcommand{\minorrev}[1]{\textcolor{black}{#1}}

\usepackage{etoolbox}
\BeforeBeginEnvironment{IEEEbiography}{\vspace{-3.7\baselineskip}}

\begin{document}

\title{Is Diversity All You Need for Scalable Robotic Manipulation?}

\author{
Modi Shi,~\IEEEmembership{Graduate Student Member,~IEEE,} Li Chen,~\IEEEmembership{Graduate Student Member,~IEEE,} Jin Chen, Yuxiang Lu,~\IEEEmembership{Graduate Student Member,~IEEE,} Chiming Liu, Guanghui Ren, Ping Luo,~\IEEEmembership{Senior Member,~IEEE}, Di Huang,~\IEEEmembership{Senior Member,~IEEE}, Maoqing Yao and Hongyang Li,~\IEEEmembership{Senior Member,~IEEE}
}

\markboth{IEEE Transactions on Robotics}%
{Shi \MakeLowercase{\textit{et al.}}: Is Diversity All You Need for Scalable Robotic Manipulation?}

\noindent
\twocolumn[{%
\renewcommand\twocolumn[1][]{#1}
\maketitle

\vspace{-36pt}
\begin{center}
    \centering
    \captionsetup{type=figure}
    \includegraphics[width=0.99\textwidth]{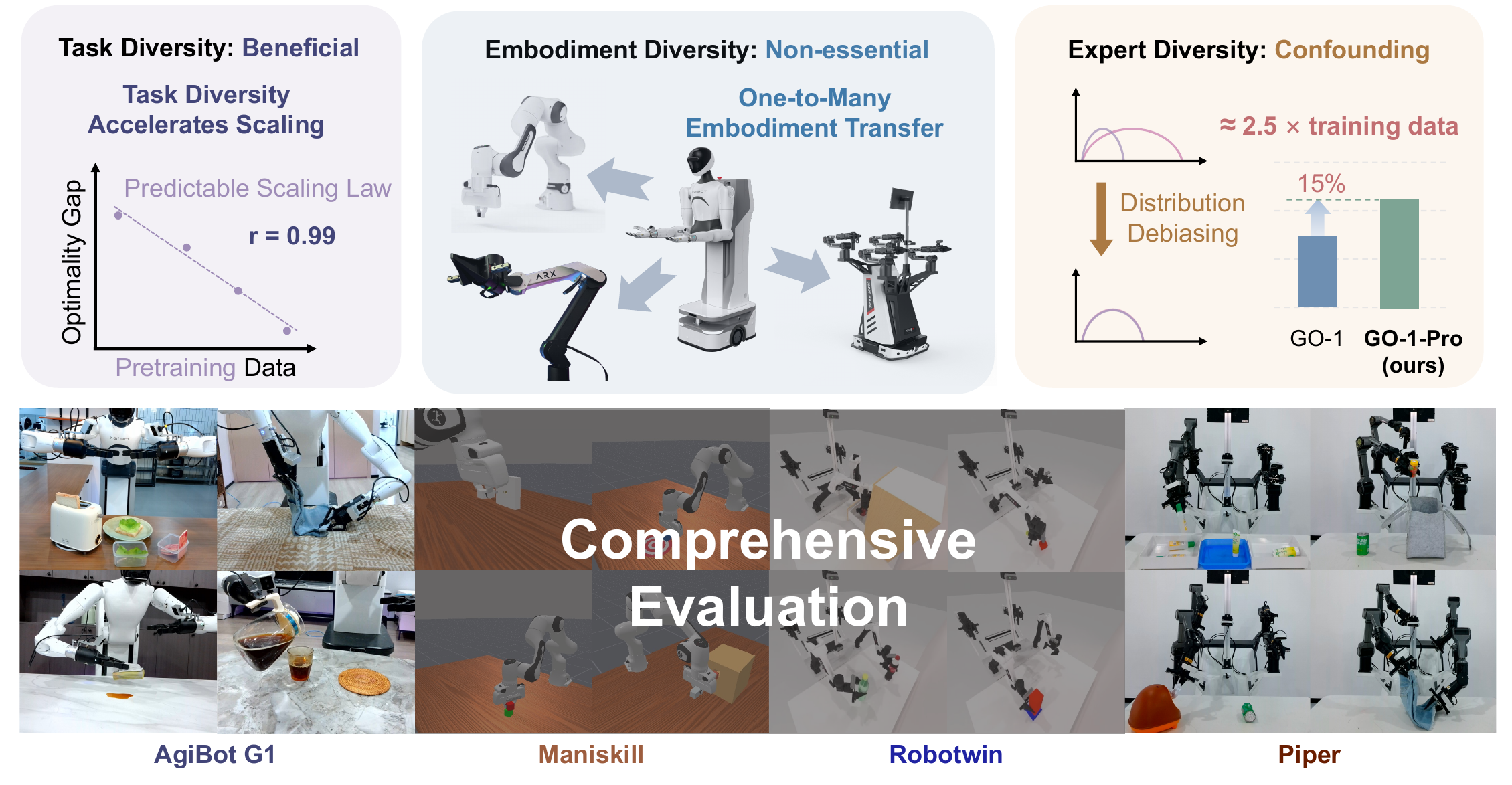}
    \vspace{-8pt}
    \captionof{figure}{
        \textbf{We investigate critical aspects of data diversity for robotic manipulation systematically, i.e., task, embodiment, and expert diversity.} Through comprehensive evaluation in simulation and the real world, we reveal key insights that challenge conventional assumptions on data scaling. (a) Task diversity benefits policy learning with predictable power-law scaling. (b) Multi-embodiment pre-training data is \minorrev{non-essential} for cross-embodiment transfer capabilities-models pre-trained on \minorrev{high-quality} single-embodiment data can efficiently adapt to different embodiments and show desirable scaling properties during finetuning, which \minorrev{can serve as a competitive alternative to large-scale multi-embodiment pre-training}. (c) Expert diversity confuses robot learning, towards which we devise a distribution debiasing method based on GO-1~\cite{go1}; the yielding \textbf{GO-1-Pro} attains superior data efficiency during both pre-training and finetuning, where it achieves substantial performance gains of 15\%, equivalent to using 2.5 times the pre-training data.
        Project page: \texttt{\url{https://github.com/OpenDriveLab/AgiBot-World}}.
    \label{fig:teaser}
    }
\end{center}
}]

\makeatletter
\def\blfootnotetext#1{%
  \begingroup
  \let\thefootnote\relax
  \let\@makefnmark\relax
  \let\Hy@footnote@currentHref\@empty
  \def\@thefnmark{}
  \@footnotetext{#1}
  \endgroup
}
\makeatother
\blfootnotetext{
This work is in part supported by the JC STEM Lab of Autonomous Intelligent Systems funded by The Hong Kong Jockey Club Charities Trust. \textit{(Correspondence author: Hongyang Li.)}\newline
\indent Modi Shi and Jin Chen are with Shanghai Innovation Institute, Shanghai, China (email: modishi@buaa.edu.cn).\newline
\indent Li Chen, Ping Luo, and Hongyang Li are with The University of Hong Kong, Hong Kong SAR, China (email: hongyang@hku.hk).\newline
\indent Yuxiang Lu, Chiming Liu, Guanghui Ren, and Maoqing Yao are with AgiBot, Shanghai, China.\newline
\indent Modi Shi and Di Huang are with Beihang University, Beijing, China.\newline
\indent Modi Shi, Li Chen, and Jin Chen contribute equally to this project.
}

\begin{abstract}
\input{text_annotated/abstract}
\end{abstract}


\section{Introduction}

\input{text_annotated/intro.tex}

\section{Related Work}
\input{text_annotated/related_work}

\section{Task Diversity}
\input{text_annotated/task_diversity}

\section{Embodiment Diversity}
\input{text_annotated/embodiment_diversity}

\section{Expert Diversity}
\input{text_annotated/expert_diversity}

\section{Conclusion and  Future Work}
\input{text_annotated/conclusion}

{
\bibliographystyle{IEEEtran}
\bibliography{bibliography_short, bibliography_custom}
}

\vspace{-\baselineskip}

\input{bio}

\section*{Appendix}
\input{text_annotated/appendix}

\end{document}

%% file: text_annotated/abstract.tex
Data scaling has driven remarkable success in foundation models 
for Natural Language Processing (NLP) and Computer Vision (CV), yet the principles of effective data scaling in robotic manipulation remain insufficiently understood. 
In this work, we investigate the nuanced role of data diversity in robot learning by 
examining three critical dimensions—task (what to do), embodiment (which robot to use), and expert (who demonstrates)—challenging the conventional intuition of ``more diverse is better''. 
Throughout extensive experiments on various robot platforms, we reveal that (1) task diversity proves more critical than per-task demonstration quantity, \change{with scene diversity playing a more important role than skill diversity for robustness and generalization under distribution shifts}; (2) multi-embodiment pre-training data is \minorrev{non-essential} for cross-embodiment transfer—models trained on high-quality single-embodiment data can efficiently transfer to different platforms, showing desirable scaling property during fine-tuning and \minorrev{its potential of replacing large-scale multi-embodiment pre-training};
and (3) expert diversity, arising from individual operational preferences and stochastic variations in human demonstrations, can be confounding to policy learning, with \change{action rate} multimodality emerging as a key contributing factor. Based on this insight, we propose a distribution debiasing method to mitigate \change{action rate} ambiguity, the yielding GO-1-Pro achieves substantial performance gains of 15\%, equivalent to using 2.5× pre-training data.
Collectively, these findings provide new perspectives and offer practical guidance on how to scale robotic manipulation datasets effectively. The code will be released.

%% file: text_annotated/intro.tex
\IEEEPARstart{R}{ecent} advances in foundation models across NLP and CV have demonstrated remarkable generalization capabilities, such as
GPT-4~\cite{gpt4}, Gemini~\cite{gemini}, and SAM2~\cite{sam2}.
A critical factor underlying these breakthroughs is systematic data scaling, where training on massive, diverse, while carefully curated datasets yields superior performance and broader applicability. Given that data scaling principles have revolutionized multiple domains, a natural question emerges: can similar data scaling approaches pave the way toward robotic foundation models?

Building on state-of-the-art Vision Language Models (VLMs)~\cite{clip,llava,blip2,internvl,qwen25vl,prismatic} and visual foundation models~\cite{vit,mae,dinov2,sam2}, the robotics community has developed several large-scale robotic models including RT-2~\cite{brohan2023rt2}, OpenVLA~\cite{kim2024openvla}, Pi-0~\cite{pi0}, RDT~\cite{rdt}, GO-1~\cite{go1}, UniVLA~\cite{univla}, and GR00T~\cite{gr00t}. Despite representing significant progress, these models are still far from genuine robotic foundation models. Their generalization capabilities remain constrained, struggling with novel objects, unfamiliar environments, new tasks, and different robot embodiments. Even minor variations in object positioning or lighting conditions can significantly compromise performance~\cite{chen2025position}. This persistent gap between robotic manipulation and advances in NLP and CV domains can be attributed to multiple factors, with the limited quantity and quality of robot datasets being a critical bottleneck. To address this data scarcity, recent efforts have focused on large-scale data collection initiatives such as Bridge Data~\cite{bridgev1, walke2023bridgedata}, DROID~\cite{droid}, Open X-Embodiment (OXE)~\cite{oxe}, and AgiBot World~\cite{go1}. However, these approaches primarily advocate a ``more is better'' philosophy, relying on brute-force collection or simple aggregation, without carefully considering what constitutes effective data for robot learning. This limitation is exemplified by OpenVLA's finding that removing DROID data actually improves model performance~\cite{kim2024openvla}. Such results raise fundamental questions about what makes a good manipulation dataset and how we should strategically scale up the datasets to maximize learning outcomes.

Some preliminary studies have attempted to address these questions by scaling data for direct Imitation Learning (IL). For instance, Lin et al.~\cite{gaoyanglaw} find a power-law relationship between policy performance and object diversity and environment diversity, while ManiBox~\cite{manibox} highlights the benefits of spatial diversity for spatial generalization. These findings naturally lead to the intuition that data diversity is universally beneficial for robotic manipulation. In this study, we systematically explore three underexplored dimensions of data diversity, e.g., task diversity, embodiment diversity, and expert diversity, and investigate data diversity in both policy pre-training and fine-tuning stages, to provide comprehensive insights into effective data scaling strategies. In the end, our investigation suggests a more complex picture. We find that the impact of diversity varies significantly across different dimensions: while some aspects of diversity are indeed critical and beneficial, others may be less important, or even confounding. 

First, we investigate how task diversity in large-scale pre-training affects downstream performance. Given that current robotic foundation models struggle to adapt to new tasks or skills, a fundamental challenge emerges regarding how models should acquire transferable knowledge. Two potential approaches exist: broad exposure to diverse tasks for general knowledge acquisition, or intensive training on focused skill sets \change{and scenarios} for specialized expertise development. To address this challenge, we construct \change{three} pre-training datasets with identical sample sizes but different task compositions: \change{one with high task diversity (episode-based sampling), one focused on target-relevant skills (task-based sampling), and one with controlled visual scene diversity (scenario-based sampling)}. Our results demonstrate that task diversity outweighs the number of demonstrations per task. \change{More importantly, we find that scene diversity plays a more critical role than skill diversity in enhancing robustness and generalization under distribution shifts.} Building on this insight, we further investigate whether model performance continues to improve with increasing training samples when task diversity is sufficiently maintained, examining the power-law scaling relationship under this condition.

Second, we explore embodiment diversity and its implications for cross-embodiment generalization. A truly foundational robotic model should be capable of adapting to different robot embodiments. While the robotics community conventionally considers that achieving this capability requires diverse embodiments in the training data, cross-embodiment training is inherently complex due to morphological and state space heterogeneity across robots~\cite{wang2024scaling,zheng2025universal,yang2024pushing,doshi2024scaling}. However, intuitively, the end-effector action spaces of robots with different configurations are fundamentally similar—robots with different morphologies can produce comparable behaviors when their end-effectors follow the same trajectory in world coordinates, suggesting that action space transformation across embodiments may be feasible. This observation leads to a critical hypothesis: models pre-trained on a single embodiment may easily transfer their learned knowledge to new robot configurations, thereby circumventing the difficulties of cross-embodiment training. To verify this hypothesis, we evaluate models trained solely on AgiBot G1 across diverse simulated and real-world platforms. Remarkably, we find that the model adapts well, even showing a more desirable scaling property during fine-tuning than models pre-trained on the OXE dataset~\cite{oxe}, which includes the test embodiment and thus has a smaller embodiment gap. Importantly, \change{we are not attempting to prove that single-embodiment pre-training is universally superior. Rather, our findings provide existence proof: \minorrev{high-quality and consistent} single-embodiment pre-training alone can achieve cross-embodiment adaptation capabilities comparable to, or even better than, models pre-trained on the massive OXE dataset. This challenges the conventional wisdom that pre-training must include the target embodiment to ensure generalization, and leads to our core conclusion: \textit{multi-embodiment pre-training is \minorrev{non-essential or not strictly required}, prioritizing high-quality, consistent data, even from a single robot, is a practical pathway that circumvents the inherent challenges of cross-embodiment training.}}

\begin{figure}[t]
\centering
\includegraphics[width=0.9\columnwidth]{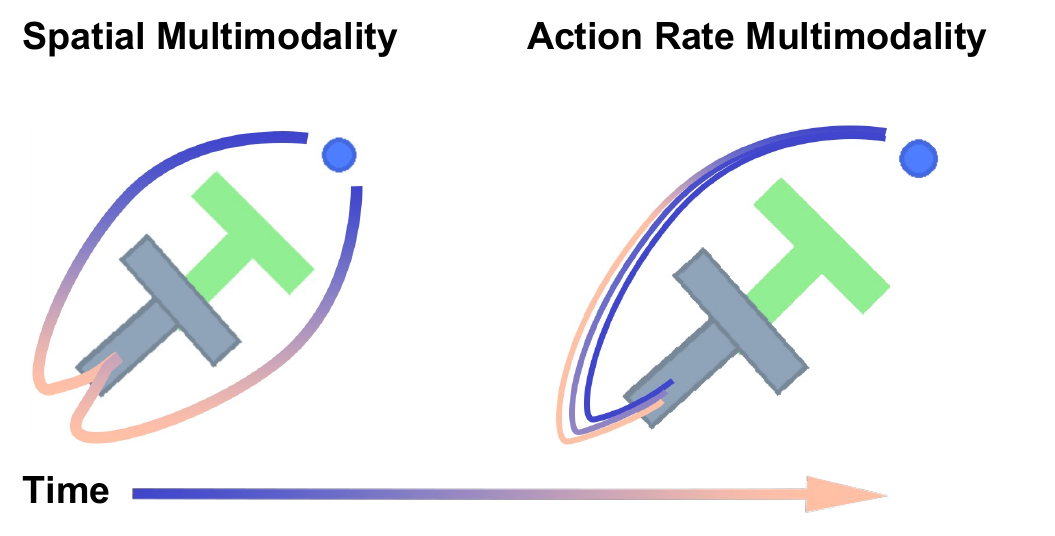}
\caption{\change{\textbf{Illustration of the multimodal expert behavior in task Push-T~\cite{chi2023diffusion}.}} The robot (blue circle) needs to move the gray T to the green target area. Expert demonstrations exhibit multimodality in both spatial and \change{action rate} dimensions: (a) Spatial multimodality arises from different trajectory choices, where the robot can approach T from either left or right sides, resulting in distinct spatial paths; (b) \change{Action rate} multimodality occurs when robots execute similar trajectory at different speeds, generating completely different demonstration profiles over time. Both spatial and action rate multimodal characteristics require models to learn these distributional properties in current action chunk-based imitation learning.}
\label{fig:multimodal-illustration}
\end{figure}

Third, we study expert diversity, an often overlooked aspect in robot learning. Expert diversity refers to the distributional variations in collected demonstrations arising from different teleoperators’ habits, skill levels, and inherent randomness. Unlike standardized NLP and CV datasets collected from the internet, robotic datasets are composed of continuous robot motion that is highly sensitive to teleoperator behaviors~\cite{li2025train}. This sensitivity results in demonstrations that, while achieving the same task, exhibit distinct distributional characteristics. As shown in Figure~\ref{fig:multimodal-illustration}, expert diversity manifests as both spatial multimodality~\cite{chi2023diffusion} in trajectory paths and \change{action rate} multimodality in execution speeds. Crucially, these two types of multimodality have fundamentally different implications for learning: spatial variations represent meaningful task strategies that should be retained, whereas \change{action rate} variations introduce undesirable noise that complicates training. To address this, we introduce \change{an action rate} model that performs distribution debiasing, specifically eliminating \change{action rate} multimodality while preserving spatial multimodality. Our experiments show that this design significantly improves model performance. These results shed light on the fundamental distinction between robotic action data and image/text data, revealing that current imitation learning approaches may be limited by overlooked data characteristics rather than insufficient model capacity or dataset scale.

We hope this consolidation report will shed new light on scalable robotic manipulation and offer practical guidelines for the research community. In summary, the contributions of this article are as follows:
\begin{enumerate}
\item \change{We demonstrate task diversity's benefits for robot learning and validate power-law scaling relationships. Crucially, scene diversity outweighs skill diversity for robustness and generalization.}
\item We show that multi-embodiment data is \minorrev{non-essential} for cross-embodiment transfer, as models pre-trained on single-embodiment data \minorrev{with high quality} can efficiently adapt to different embodiments with desirable scaling property during fine-tuning, serving as effective alternatives to multi-embodiment pre-trained models.
\item We discover that expert diversity could be confounding to the imitation learning process and demonstrate that targeted elimination of \change{action rate} multimodality can significantly improve model performance.
\end{enumerate}

%% file: text_annotated/related_work.tex
\subsection{Scalable Manipulation Policy}
Recently, the success of foundation models in CV~\cite{vit,clip,mae,dinov2} and NLP~\cite{bert,llama,gpt3,minigpt4} has motivated research into establishing manipulation foundation models~\cite{li2023roboflamingo,driess2023palm,wen2025tinyvla,wen2025dexvla,kim2024openvla,pi0,gr00t} through IL, aiming to leverage large-scale pre-training data to enable models to absorb rich prior knowledge. Early studies like RT-1~\cite{brohan2022rt1} and Octo~\cite{team2024octo} employ transformer-based policies to learn generalizable manipulation knowledge from diverse heterogeneous data. 
Works like RT-2~\cite{brohan2023rt2} and OpenVLA~\cite{kim2024openvla} utilize advanced VLMs to process camera inputs and human instructions, which have been extensively trained on web-scale image-text pairs and possess substantial world knowledge. However, these methods directly integrate VLMs to generate low-level robotic actions, which may not be optimal for achieving generalization across different embodiments and skills. Some works utilize large-scale human manipulation videos~\cite{something2something,ego4d} to pre-train models that learn inverse dynamics by predicting latent action representations~\cite{genie,lapa,igor}, or forward dynamics by predicting future changes~\cite{gr1,gr2,unipi,mpi,clover}, thus enhancing cross-embodiment and cross-skill generalization. Other works focus on designing policy architectures with stronger capabilities for modeling multimodal distributions. Diffusion Policy~\cite{chi2023diffusion} incorporates a diffusion module to adapt to trajectory multimodality. RDT~\cite{rdt} employs DiT~\cite{dit} for improved pre-training on heterogeneous multi-robot datasets~\cite{oxe} and dual-arm trajectories. In our work, we conduct experiments based on GO-1~\cite{go1} and RDT~\cite{rdt} to ensure generalization potential across tasks and embodiments.

\subsection{Large-scale Manipulation Dataset}
Robotic manipulation is experiencing a significant transformation toward scaling up data~\cite{bridgev1,walke2023bridgedata,droid,rh20t,ario,brohan2022rt1,robomind,wu2025freetacman}, seeking to enable models to develop general manipulation capabilities. The Open X-Embodiment dataset~\cite{oxe} exemplifies this effort by consolidating multiple datasets across 22 different embodiments and various camera configurations, reaching a significant scale of 2.4 million trajectories. DROID~\cite{droid} emphasizes increasing data diversity by covering various tasks, objects, scenes, camera viewpoints, and interaction locations, collecting data across 564 scenes in 52 real-world buildings. AgiBot World~\cite{go1} has compiled over 1 million high-quality bimanual trajectories through a unified embodiment and camera configuration, skilled teleoperators and rigorous verification protocols. Additionally, other works explore learning robotic manipulation knowledge from human manipulation videos~\cite{ego4d,something2something}, leveraging the abundance of such videos to address the scarcity of robotic data. In this work, we conduct a more comprehensive analysis of manipulation data diversity, aiming to provide deeper insights into constructing large-scale manipulation datasets of high quality.

\subsection{Crucial Dimensions Regarding Data Distribution}
Scaling robot datasets requires strategic approaches beyond simply expanding data volume. Recent research has demonstrated how data diversity enhances model generalization~\cite{gaoyanglaw,pan2025ams,team2025gemini,kareer2025emergence}. Lin et al.~\cite{gaoyanglaw} show that incorporating datasets with diverse environments and objects significantly improves model generalization, establishing a power-law relationship between policy performance and data diversity. Manibox~\cite{manibox} highlights spatial diversity benefits, demonstrating that varied spatial layouts improve spatial generalization in manipulation tasks. Saxena et al.~\cite{saxena2025matters} identify camera viewpoints and spatial arrangements as crucial dimensions for collection diversity and retrieval alignment. Hejna et al.~\cite{hejna2408re} propose automatic curation of large-scale robotics datasets using group distributionally robust optimization, enabling efficient utilization of heterogeneous data for imitation learning. Similar findings exist in autonomous driving. Zheng et al.~\cite{zheng2024preliminary} collect data from various driving scenarios and behaviors, discovering that driving model performance exhibits a power-law relationship with scenario and behavior diversity. Our research challenges this conventional view, suggesting that not all diversity forms are equally beneficial. While diversity in environments and objects proves crucial, other diversity types may be less important or even confounding.

\medskip
In the following sections, we address three critical aspects of diversity in detail: task diversity (Section~\ref{sec:task-diversity}), embodiment diversity (Section~\ref{sec:embodiment diversity}), and expert diversity (Section~\ref{sec:expert diversity}).

%% file: text_annotated/task_diversity.tex
\label{sec:task-diversity}
Recent studies have explored scaling properties, such as \cite{gaoyanglaw} on environment diversity, \minorrev{i.e., variations in lighting conditions, distractor objects, background changes,} \change{and RT-1~\cite{brohan2022rt1} on skill diversity}, \minorrev{which corresponds to atomic skills or verbs in task instructions, excluding the targeted objects.} Meanwhile, these analyses were limited to single-task scenarios \change{or significantly smaller scales (e.g., RT-1 utilized only six skills and $\sim$10\% of our data volume).} In contrast, we investigate how \change{different types of task diversity in} large-scale pre-training datasets affect downstream task performance, specifically examining the trade-off between \change{skill diversity (task coverage) and scenario diversity (environmental variations)}. \change{We decompose task diversity into these two fundamental components and analyze their relative importance for in-domain performance versus generalization under distribution shifts.}

\subsection{Experiment Design}
\label{subsec:exp-setting}

Our experiments employ GO-1~\cite{go1} as the policy architecture, which excels at extracting task-agnostic latent actions and generalizing across diverse tasks. We leverage AgiBot World~\cite{go1}, a large-scale robotic learning dataset containing over 1 million trajectories across more than 100 real-world scenarios. This dataset offers unique advantages, as all data is collected using a single robot platform, AgiBot G1, which eliminates cross-embodiment variables while ensuring high data quality through standardized collection protocols. Our training follows a two-phase process: first pre-training on the large-scale manipulation dataset, followed by fine-tuning on target evaluation tasks. \change{To isolate the effects of skill diversity and scene diversity, we construct three pre-training datasets with identical data quantity (10\%) but different sampling strategies: scenario-based sampling (limited scenario diversity), task-based sampling (limited skill diversity), and episode-based sampling (full spectrum of task variety). We then fine-tune all models on identical evaluation task data to compare how different diversity compositions affect downstream performance.}

Our evaluation encompasses four challenging tasks: Wipe Table (contact-rich cleaning), Fold Shorts (deformable object manipulation), Pour Water (fine-grained pouring), and Make Sandwich (long-horizon assembly). Each task is evaluated across three scenarios: an in-domain scenario, an object-environment generalization scenario, and a visual distraction scenario. We conduct ten trials per scenario with position perturbations under consistent indoor lighting conditions, ensuring identical evaluation settings across all models. For evaluation, we use normalized scores to record the performance of each trial. We establish specific scoring criteria for each action within every task, with action scores categorized into three levels: 1, 0.5, and 0. The evaluation score for each trial corresponds to the average of all action scores, where a score of 1 indicates perfect completion of all actions, and fractional scores represent partial success. Detailed scoring criteria can be found in Appendix.

\subsection{Task Diversity for Robotic Manipulation Pre-training}
\label{sec:task-diversity-B}

When applying a model to a specific downstream task out of the pre-training domain, a fundamental consideration emerges regarding whether to construct a pre-training dataset with the richest possible diversity, or to utilize fewer tasks but with potentially higher relevance to the target downstream task. To explore this trade-off, we design \change{three} pre-training datasets with distinct \change{diversity characteristics} while controlling for other factors\change{, allowing us to decompose task diversity into two key components: skill diversity and scene diversity}.

We leverage the Agibot-World Beta dataset, one of the most comprehensive robotic datasets, as our source for constructing these contrasting pre-training datasets. This allows us to maintain consistency in other aspects of the data, such as the included robot embodiments, while varying only the \change{diversity dimensions}. We employ \change{three} sampling strategies to create datasets with different diversity but identical sizes\change{:}

\begin{itemize}
    \item \change{\textbf{Scenario-based sampling (10\% Scenario):} Randomly samples 10\% of scenarios from the dataset, where each scenario represents a specific environmental setup (e.g., kitchen counter, dining table). This strategy provides limited diversity in scenes.}
    
    \item \textbf{Task-based sampling (10\% Task):} \change{Manually selects} 10\% of tasks that are most relevant to our target downstream tasks. \change{This results in a dataset with high scenario coverage but limited skill diversity.}
    
    \item \textbf{Episode-based sampling (10\% Episode):} Randomly samples 10\% of episodes from each task in the original dataset, preserving the full spectrum of task variety. \change{This strategy captures both skill diversity across different tasks and scene diversity within each task, as episodes naturally include varied object configurations, lighting conditions, and spatial arrangements.}
\end{itemize}

\begin{figure}[t]
\centering
\includegraphics[width=\columnwidth]{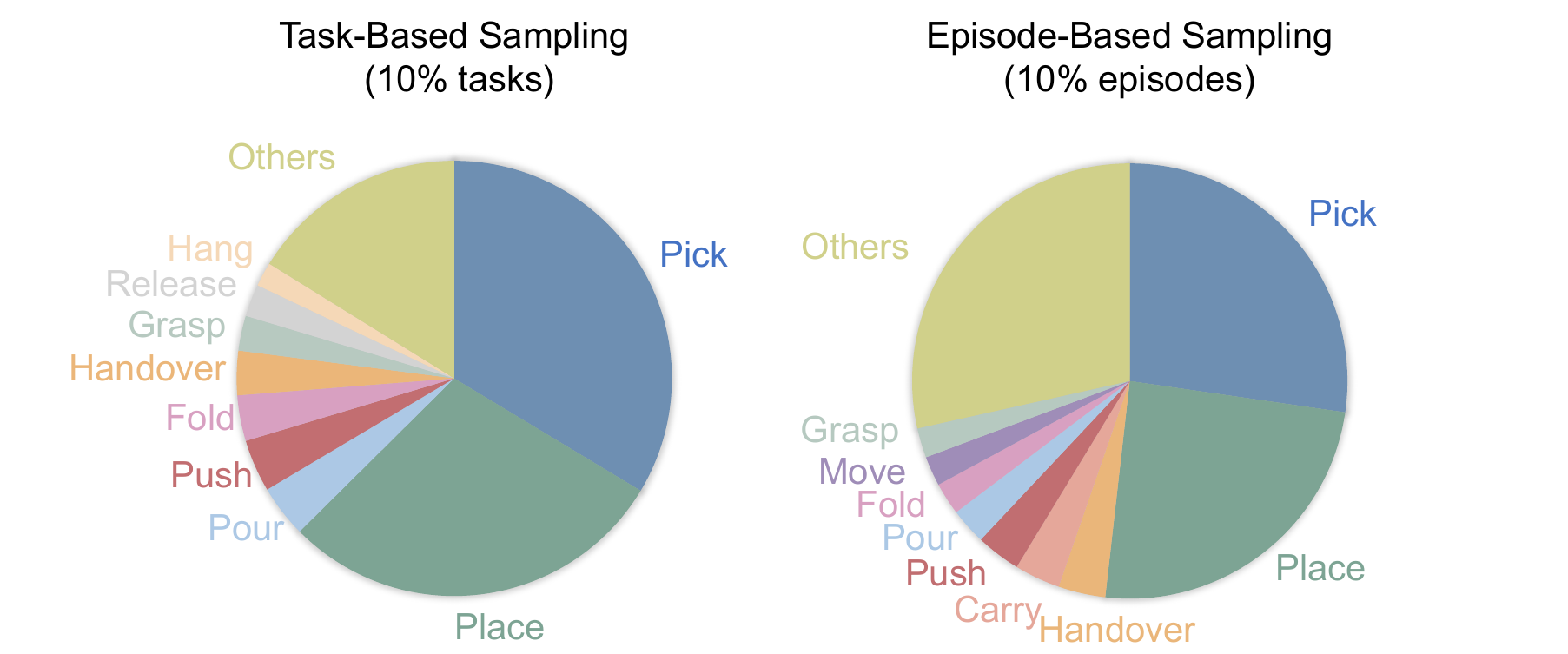}
\caption{\change{\textbf{Distribution of atomic skills in two pre-training datasets.} Task-based sampling (10\% tasks) shows lower skill diversity but concentrates on the most commonly used skills, while episode-based sampling (10\% episodes) demonstrates a more balanced distribution.}}
\label{fig:skill-distribution}
\end{figure}

Our criterion for selecting relevant tasks is based on the inclusion of atomic skills required for task completion. Our evaluation tasks encompass five common atomic skills: pick, place, grasp, pour, and fold. In Figure~\ref{fig:skill-distribution}, we present the distribution of atomic skills in both pre-training datasets, the episode-sampled dataset exhibits significantly greater task and skill diversity, yet consequently contains fewer episodes (59.2\% vs. 71.1\%) corresponding to the specific atomic skills needed for the target evaluation tasks.

\begin{figure*}[t]
\centering
\includegraphics[width=.9 \textwidth]{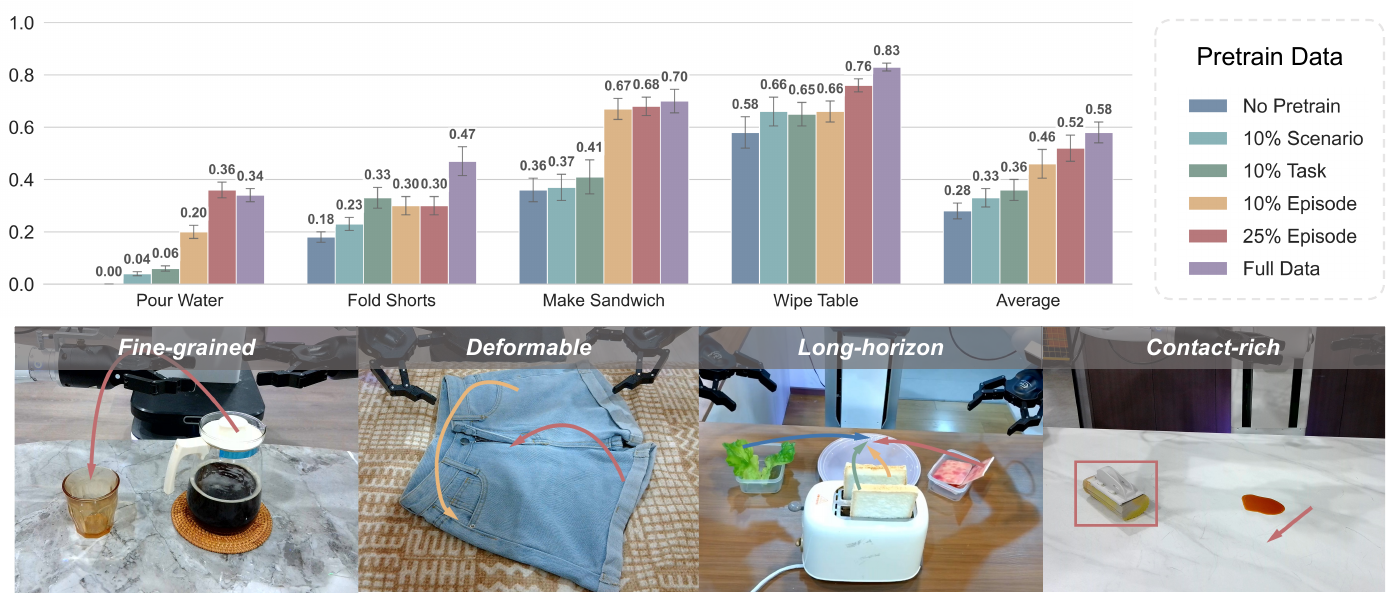}
\caption{\change{\textbf{Real-robot evaluation of GO-1~\cite{go1} on four 
challenging tasks subsequent to pre-training on different datasets.} Error bars show standard errors. The tasks assess fine-grained manipulation, deformable object handling, long-horizon planning, and contact-rich interactions respectively. Results show that episode-based sampling (10\% Episode) outperforms task-based sampling (10\% Task) by 0.1 in average score with the same data amount, and performance improves consistently with increased pre-training data while ensuring sufficient task diversity.}}
\label{fig:pre-train-his}
\end{figure*}

\begin{figure}[htbp]
\centering
\includegraphics[width=\columnwidth]{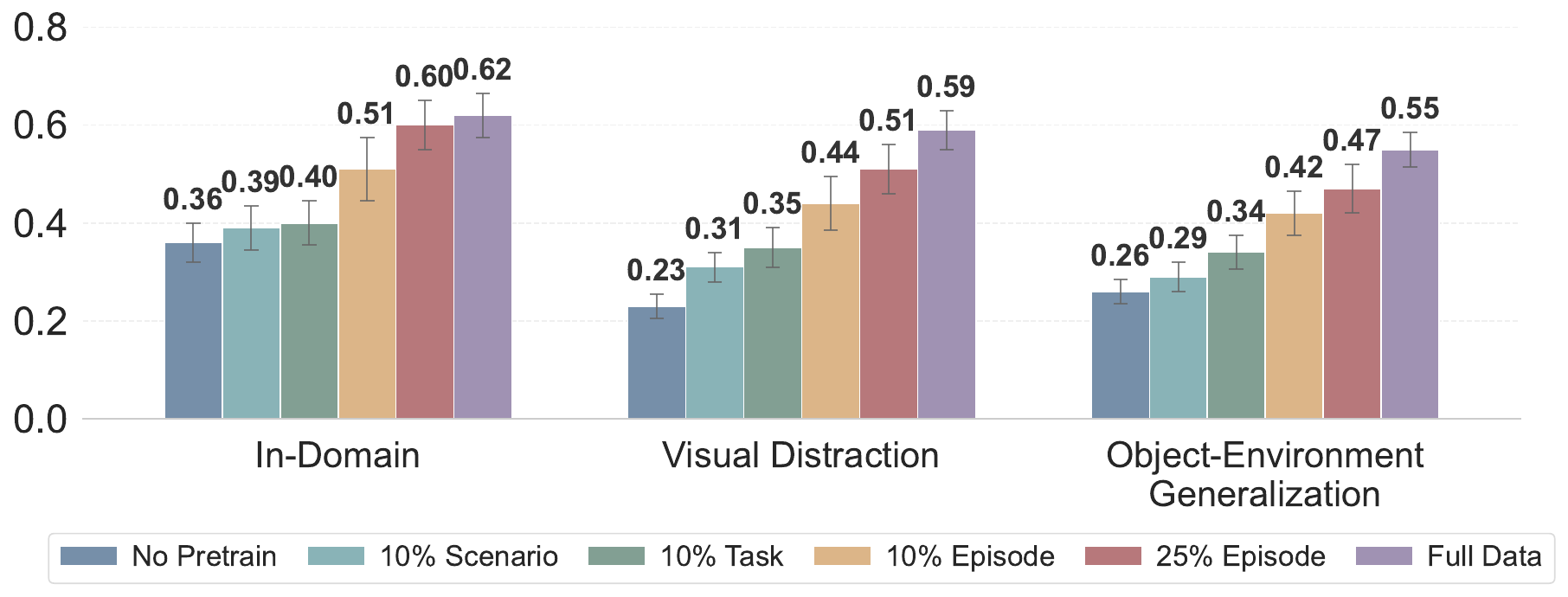}
\caption{\change{\textbf{Generalization performance across different pretraining data.}  Error bars show standard errors. The model is evaluated under three generalization scenarios: In-Domain, Visual Distraction, and Object-Environment Generalization. Episode-based sampling (10\% Episode) consistently outperforms task-based sampling (10\% Task) and scenario-based sampling (10\% Scenario). Performance improves with increased pretraining data, demonstrating the importance of both data scale and diversity for robust generalization.
}}
\label{fig:pre-train-gen-his}
\end{figure}

The experimental results in Figure~\ref{fig:pre-train-his} demonstrate that \change{data diversity significantly impacts downstream performance. At 10\% data scale, scenario-based sampling achieves 0.33 average performance, task-based sampling improves to 0.36, while episode-based sampling reaches 0.46}. The episode-based sampling approach achieves an average performance improvement of 0.1 compared to task-based sampling, with the most significant gains observed in tasks requiring higher semantic and spatial understanding, such as Make Sandwich and Pour Water. This finding corroborates the conclusions in~\cite{gaoyanglaw} that, given fixed \change{data quantity, enhancing diversity provides greater benefits than simply increasing data volume}. 

\change{\smallskip
To understand how skill diversity and scene diversity affect model capabilities under distribution shifts, we evaluate all pre-training strategies across three conditions: in-domain evaluation, visual distraction robustness, and object-environment generalization. As shown in Figure~\ref{fig:pre-train-gen-his}, in-domain performance shows episode-based sampling outperforming task-based and scenario-based by comparable margins. However, under distribution shifts, scene diversity demonstrates greater impact. For visual distraction, episode-based exceeds scenario-based by 0.13 compared to only 0.09 over task-based. In object-environment generalization, this pattern intensifies: episode-based surpasses scenario-based by 0.13 versus 0.08 over task-based.}

\change{These results reveal that \textbf{scene diversity is more critical than skill diversity for robustness and generalization}. While both diversity types provide similar in-domain benefits, scene diversity advantages consistently exceed skill diversity advantages under distribution shifts with statistically significant differences (\(p < 0.05\)). The continued gains from data scaling highlight its capacity to enhance robustness and generalization in real-world deployment.}

\subsection{Pre-training Data Scaling Law}

\begin{figure}[t]
\centering
\includegraphics[width=\columnwidth]{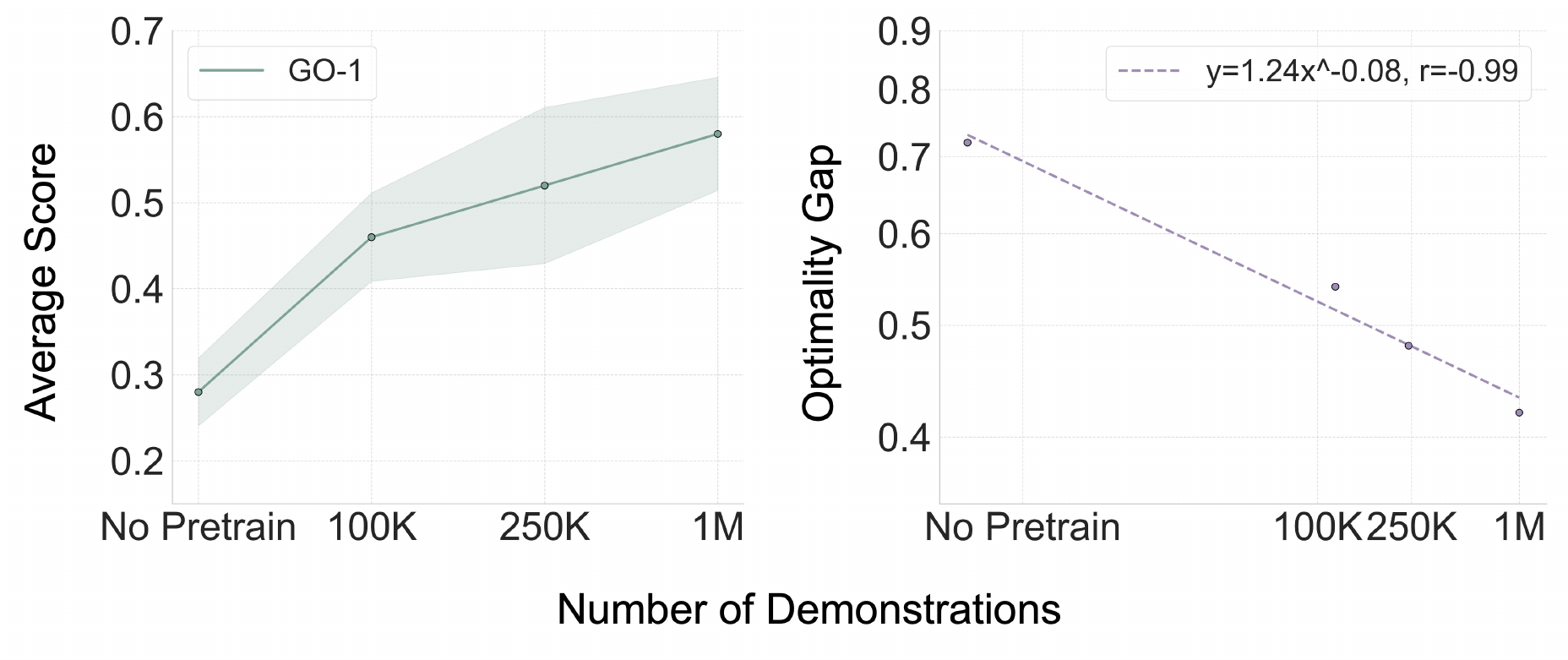}
\caption{\textbf{Performance scales with pre-training data size while maintaining adequate task diversity, following a predictable power-law relationship.} \textbf{Left}: GO-1 performance scales with pre-training data size. \textbf{Right}: Power-law relationship between pre-training data size and model performance. The dashed line represents a power-law fit with equation $y=1.24x^{-0.08}$ and correlation coefficient $r=-0.99$, indicating a strong adherence to power-law scaling with pre-training data volume.}
\label{fig:pre-train-power-law}
\end{figure}

The Agibot World~\cite{go1} encompasses 217 common daily tasks and 87 frequently used skills. Building upon the conclusion that task diversity benefits robotic learning, we further investigate the relationship between data quantity and performance when scaling up pre-training data while maintaining dataset diversity. The left panel in Figure~\ref{fig:pre-train-power-law} demonstrates consistent performance improvements across different pre-training data scales, with GO-1 average scores increasing from 0.28 (No pre-training) to 0.47 (100K demonstrations), 0.53 (250K demonstrations), and reaching 0.58 (1M demonstrations). 

To further explore this relationship, we fit the data using a power law curve $Y = \beta \cdot X^\alpha$, where $Y$ represents the optimality gap, defined as the deviation from the maximum score (i.e., $1 - \text{Normalized Score}$), and $X$ represents the number of demonstrations~\cite{gaoyanglaw}. Since the no pre-train case corresponds to zero pre-training data, which cannot be fitted in the scaling law curve, we use the fine-tuning data quantity to replace the number of demonstrations for fitting purposes. The experimental results in Figure~\ref{fig:pre-train-power-law} reveal a clear power-law relationship between model performance and pre-training data, with a Pearson correlation coefficient reaching $-0.99$. This finding suggests that, under the condition of adequate task diversity, robotic learning can achieve systematic performance gains through increased data scale, providing a clear path for developing more capable robotic systems through data scaling.

%% file: text_annotated/embodiment_diversity.tex
\label{sec:embodiment diversity}
Cross-embodiment learning faces significant challenges due to morphological and state space heterogeneity across robot platforms. However, it remains unclear whether pre-training datasets must include multi-embodiment data to achieve effective cross-embodiment transfer. In this section, we investigate whether single-embodiment pre-training—thereby avoiding cross-embodiment training complexities—can still yield models with cross-embodiment capabilities.

\subsection{Experiment Design}
\label{sec:embodiment diversity-exp design}
To address the challenges of cross-embodiment training, we explore whether single-embodiment pre-training can still enable effective cross-embodiment transfer by utilizing the large-scale single-embodiment dataset Agibot World (1M trajectories from AgiBot G1) for pre-training and systematically evaluating the resulting model's cross-embodiment generalization. \change{As a reference point, we include RDT~\cite{rdt} pre-trained on OXE~\cite{oxe}, a widely recognized architecture specifically designed for multi-embodiment learning with high-quality, community-validated open-source checkpoints. \minorrev{Note that OXE exhibits extensive atomic skills compared to Agibot World, rich in task diversity mentioned in Section~\ref{sec:task-diversity}}. This ensures the multi-embodiment baseline represents a strong, state-of-the-art standard, allowing performance differences to be strictly attributed to data diversity rather than implementation quality.}

\change{
We evaluate cross-embodiment transfer across 3 distinct benchmarks, ManiSkill (Franka arm), RoboTwin (Arx arm), and real-world Agilex (Cobot Magic), which all exhibit substantial differences from the AgiBot G1 used in pre-training. Regarding \textbf{action spaces}: Agibot G1 uses dual-arm 12-DOF delta end-effector + 2-DOF grippers; ManiSkill's Franka uses single-arm 7-DOF absolute joint + 1-DOF gripper; RoboTwin's Arx and Agilex's Cobot Magic both use dual-arm 12-DOF absolute joint + 2-DOF grippers, with Cobot Magic adding 2-DOF mobile base control. For \textbf{camera configurations}: Agibot G1 employs 1 head-view + 2 wrist-view cameras; ManiSkill uses only 1 third-person view; RoboTwin and Agilex use 2 wrist-view + 1 head-view cameras. While RoboTwin and Agilex share the same camera configuration type as Agibot G1, camera mounting positions, viewing angles, and fields of view differ significantly due to different robot morphologies and workspace setups. Moreover, simulated visual rendering in ManiSkill~\cite{maniskill} and RoboTwin~\cite{robotwin} introduces a substantial sim-to-real gap compared to real-world Agibot World data.}

\change{Importantly, all evaluation settings (absolute joint control, camera configurations) are already present in OXE's training data (Franka robots, dual-arm systems, diverse cameras), giving RDT-OXE prior exposure to these exact embodiments. In contrast, RDT-AWB must transfer from a completely different embodiment. This means the comparison is not entirely fair, where, in fact, RDT-OXE has an inherent advantage through embodiment alignment with evaluation benchmarks.
}

Performance is measured by the average success rates for simulation tasks and the average scores for real-world tasks. ManiSkill includes 5 tasks (PegInsertionSide, PickCube, StackCube, PlugCharger, PushCube), while RoboTwin includes 4 tasks (BlockHammerBeat, BlocksStack, ContainerPlace, DualBottlesPick). Each simulation task involves 25 rollouts across 10 random seeds. Real-world evaluation covers \change{5} tasks: Package Product, Fold Shorts, Clean Trash, Industrial Sorting, and \change{Push Chairs}.

\subsection{One-to-Many Embodiment Transfer Evaluation}
\label{sec:embodiment diversity-exp results}
\begin{figure}[t]
\centering
\includegraphics[width=0.99\columnwidth]{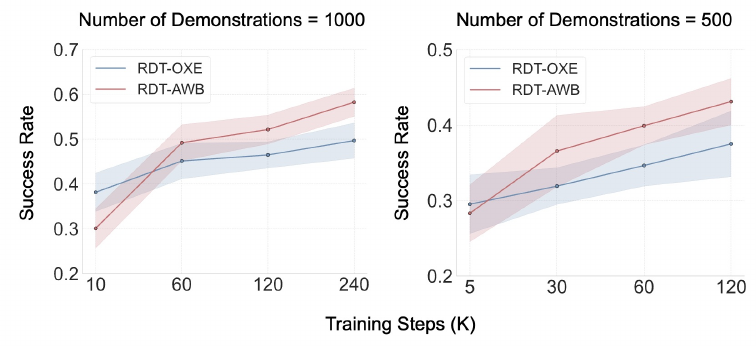}
\caption{\textbf{Cross-embodiment adaptation to Franka arm in ManiSkill with varying training steps}. \textbf{Left:} Performance vs. training steps with 1000 fine-tuning episodes per task. \textbf{Right:} Performance vs. training steps with 500 fine-tuning episodes per task.}
\label{fig:mani-step-scale}
\end{figure}

We fine-tune two pre-trained models on ManiSkill: RDT-OXE (pre-trained on OXE) and RDT-AWB (pre-trained on Agibot World beta). OXE includes the Franka robot embodiment and ManiSkill data, while Agibot World uses a completely different robot embodiment. Conventional wisdom suggests that RDT-AWB should underperform RDT-OXE or require substantially more fine-tuning due to the cross-embodiment gap. However, our results in Figure~\ref{fig:mani-step-scale} demonstrate that single-embodiment pre-training can also achieve effective cross-embodiment capabilities. While RDT-OXE converges faster initially and slightly outperforms RDT-AWB in early stages, RDT-AWB achieves effective cross-embodiment adaptation and surpasses RDT-OXE without requiring extensive fine-tuning data or training steps.

\begin{figure}[t]
\centering
\includegraphics[width=0.99\columnwidth]{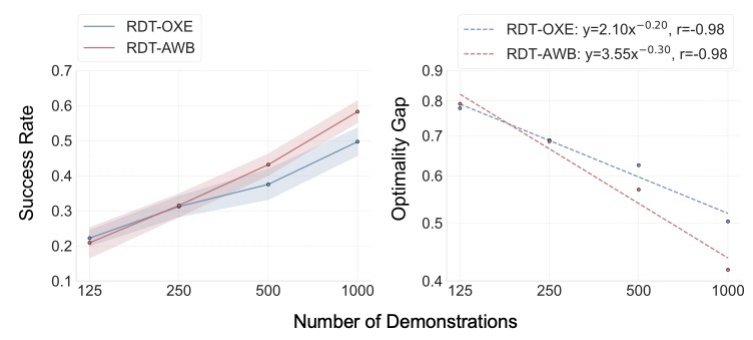}
\caption{\textbf{Cross-embodiment adaptation to Franka arm in ManiSkill with varying training data sizes}. \textbf{Left:} Performance vs. number of demonstrations per task in fine-tuning data. \textbf{Right:} Power-law relationship between downstream performance and fine-tuning data size.}
\label{fig:mani-size-scale}
\end{figure}

As shown in Figure~\ref{fig:mani-size-scale}, with 125 samples per task, RDT-OXE performs slightly better. At 250 samples, RDT-AWB matches RDT-OXE. With more data, RDT-AWB surpasses RDT-OXE\footnote{Liu et al.~\cite{rdt} report RDT-OXE's fine-tuning performance on ManiSkill as 53.6\%. Our evaluation results differ, possibly due to varying training configurations or inconsistent inference random seeds. Importantly, our RDT-OXE and RDT-AWB evaluations use identical training and inference setups.}, with the gap increasing proportionally, exhibiting a power-law relationship. Similarly, Figure~\ref{fig:mani-step-scale} shows RDT-OXE performing better with fewer steps (around 10,000), but RDT-AWB surpasses RDT-OXE as training steps increase. These results provide compelling evidence that single-embodiment pre-training can develop robust cross-embodiment transfer capabilities while circumventing the complexities inherent in multi-embodiment training.

\begin{figure}[t]
\centering
\includegraphics[width=0.99\columnwidth]{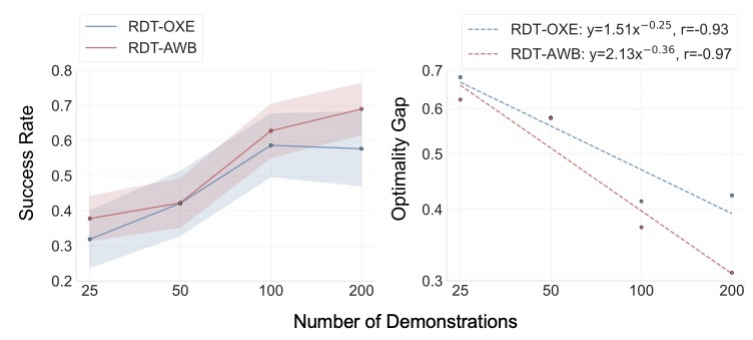}
\caption{\textbf{How the model crosses the embodiment gap to Arx in RoboTwin as data sizes increase.} \textbf{Left:} Performance vs. number of demonstrations per task in fine-tuning data. \textbf{Right:} Power-law relationship between downstream performance and fine-tuning data size.}
\label{fig:twin-size-scale}
\end{figure}

To more comprehensively validate our conclusions, we further compare RDT-OXE and RDT-AWB in both the RoboTwin~\cite{robotwin} simulation environment (using Arx) and the real-world Agilex environment (using Cobot Magic), testing cross-embodiment adaptation across multiple platforms. Following the same experimental protocol as in Figure~\ref{fig:mani-size-scale}, we conduct experiments in RoboTwin to compare model performance under varying fine-tuning data sizes and analyze the power-law relationship between performance and data size, as illustrated in Figure~\ref{fig:twin-size-scale}. The results demonstrate that RDT-AWB achieves performance comparable to that of RDT-OXE with minimal fine-tuning data, successfully adapting to Arx. Additionally, we compare the fine-tuning performance of RDT-OXE and RDT-AWB in the real-world Agilex environment using identical data sizes (100 demonstrations per task), as presented in Table~\ref{tab:agilex}. RDT-AWB achieves superior performance compared to RDT-OXE on \change{4 of 5} tasks, indicating effective adaptation to the real-world Cobot Magic with limited fine-tuning requirements.

\change{Collectively, these results across simulated and real-world environments substantiate our core argument: \textbf{multi-embodiment pre-training is \minorrev{non-essential, not mandatory}, for effective cross-embodiment transfer.} While RDT-OXE benefits from prior exposure to the evaluation embodiments, RDT-AWB's superior performance demonstrates that the target embodiment need not be present during pre-training to achieve strong generalization. We do not assert that single-embodiment pre-training is universally superior; rather, these findings highlight that prioritizing high-quality data, even from a single source, offers a viable and efficient alternative to the complexities of scaling multi-embodiment datasets. This challenges the prevailing assumption that embodiment diversity is a prerequisite for generalizable policy learning.}

\begin{table*}[t]
\centering
\caption{\change{\textbf{Performance in the real world Agilex environment.}} The performance of RDT-OXE and RDT-AWB is compared to assess RDT-AWB's ability to transfer to the downstream Cobot Magic robot.}
\begin{tabular}{c!{\vrule}ccccc!{\vrule}c}
\toprule[1pt]
Model   & Package Product & Fold Shorts & Clean Trash & Industrial Sorting & \change{Push Chairs} & Average Score \\ 
\midrule
RDT-OXE & 0.40$\pm$0.05           & \textbf{0.65}$\pm$0.04       & 0.33$\pm$0.02       & 0.23$\pm$0.03              & \change{0.27$\pm$0.03} &
0.38$\pm$0.03    \\
RDT-AWB & \textbf{0.57$\pm$0.06}           & 0.48$\pm$0.05       & \textbf{0.47$\pm$0.05}       & \textbf{0.27$\pm$0.02}       & \change{\textbf{0.35$\pm$0.03}}          & \textbf{0.43$\pm$0.04}    \\
\bottomrule[1pt]
\end{tabular}
\label{tab:agilex}
\end{table*}

%% file: text_annotated/expert_diversity.tex
\label{sec:expert diversity}
During the data collection process, different human demonstrators exhibit distinct collection habits and inherent randomness in their execution, leading to diverse and complex data distributions with varying trajectory patterns. While some variations in the data distribution represent meaningful multimodal spatial distributions that capture legitimate alternative approaches~\cite{chi2023diffusion}, others constitute distribution bias that merely increases training difficulty without providing valuable information. In this section, we propose a distribution debiasing method to eliminate bias in the \change{action rate} dimension (i.e., speed of the movement), thus enhancing learning efficiency and overall model performance. The experiment setting is the same as Section~\ref{subsec:exp-setting}.

\begin{figure}[t]
\centering
\includegraphics[width=0.9\columnwidth]{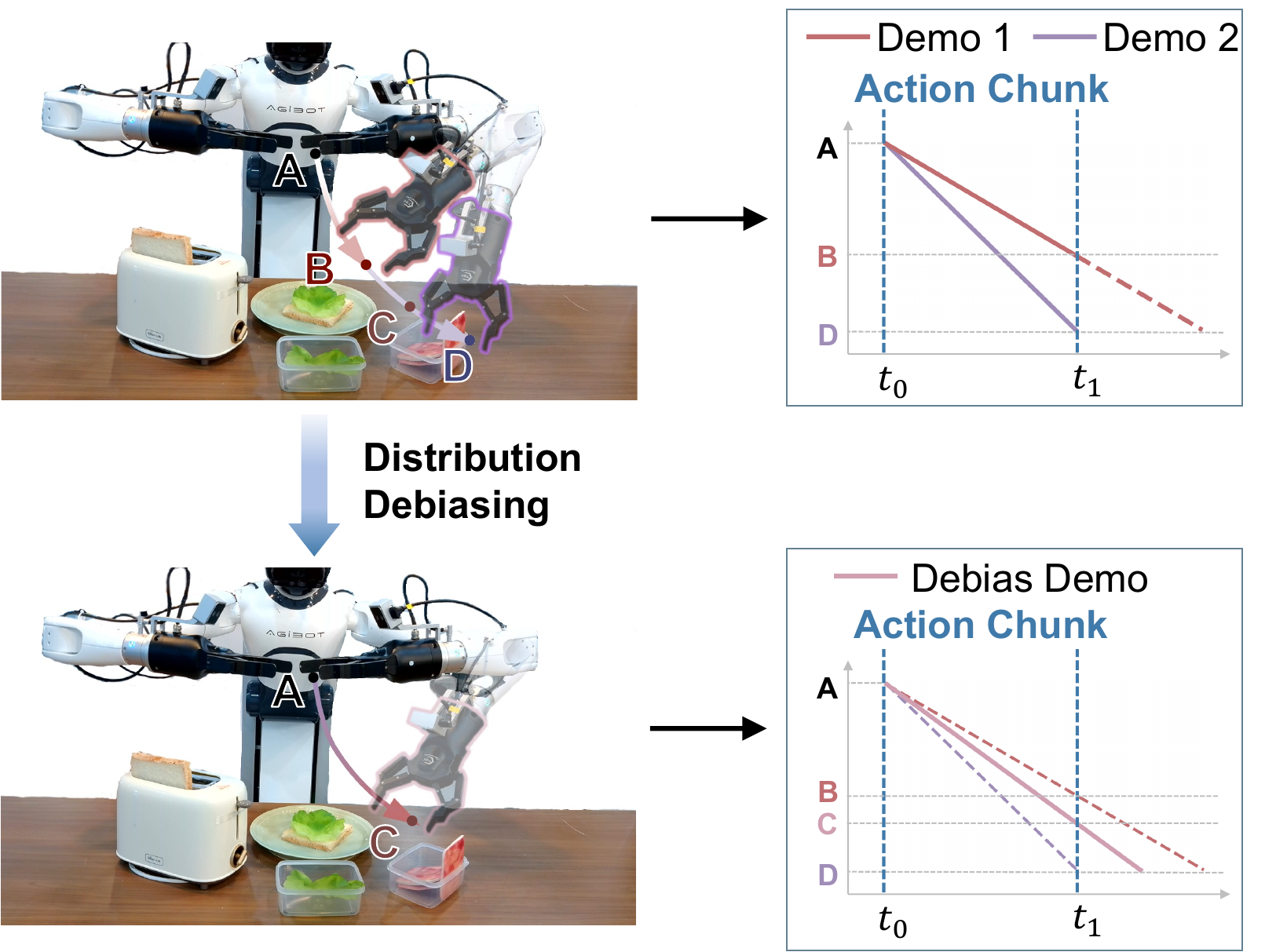}
\caption{\textbf{Illustration of distribution debiasing. Top:} Two demonstrations (Demo 1 \& 2) follow the same trajectory (A→D) but have different \change{action rates}, resulting in distinct action chunks within the same time window (A→B vs A→D). \textbf{Bottom:} After \change{action rate}-based distribution debiasing, both demonstrations are normalized to similar action chunks (A→C), reducing \change{action rate} ambiguity and facilitating model learning.
}
\label{fig:action-rate-illustration}
\end{figure}

\begin{figure}[t]
\centering
\includegraphics[width=0.9\columnwidth]{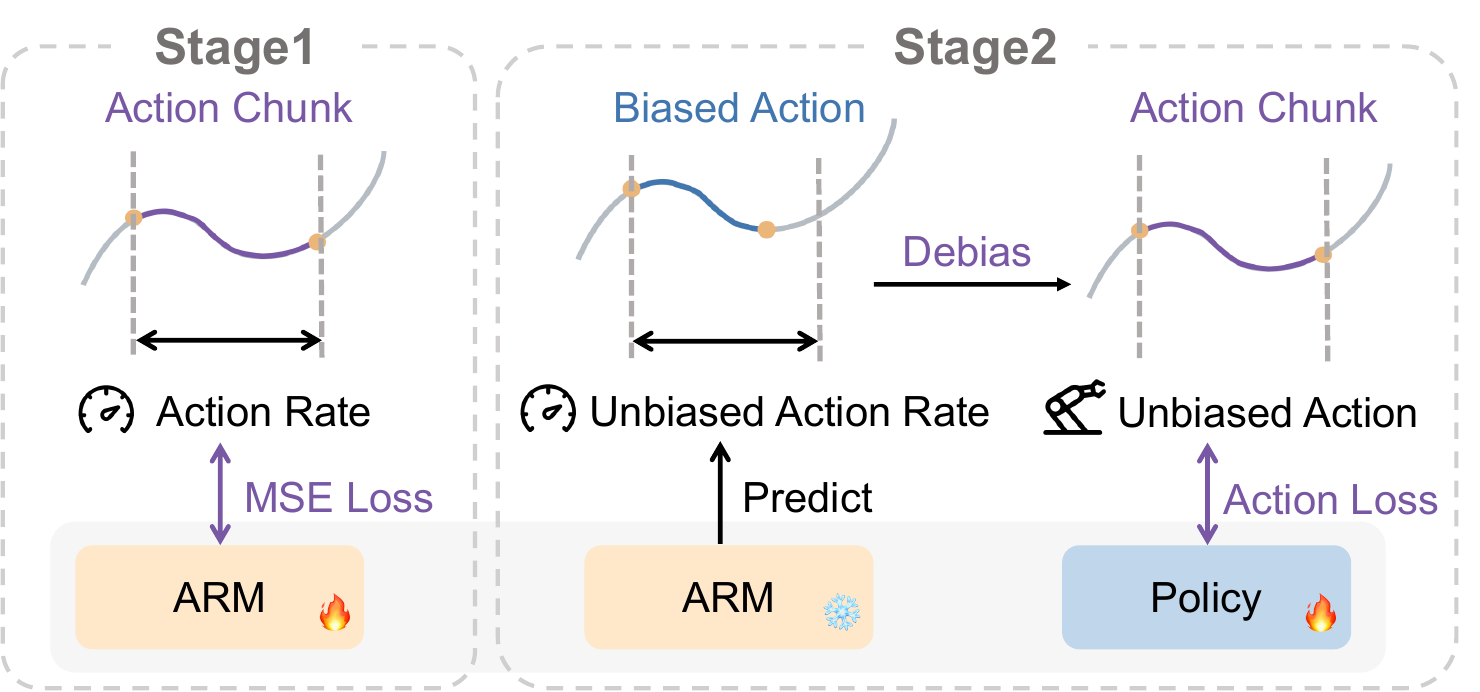}
\caption{\change{\textbf{Two-stage distribution debiasing framework using \change{ARM}. Stage 1:}} \change{ARM} is trained to predict \change{action rate} from action chunks using MSE loss, learning the expected \change{action rate} for each input from \change{action rate}-biased training data. \textbf{Stage 2:} During policy training, \change{ARM} first predicts the unbiased \change{action rate} for each training sample, which is then used to transform the original actions into unbiased actions. The policy is subsequently trained using these unbiased actions as supervision targets, effectively simplifying the distribution complexity.}
\label{fig:go-1-pro}
\end{figure}

\subsection{Distribution Debiasing}
\label{sec:deistribution-biasing}

\change{Formally, we model a demonstration trajectory as a temporal sequence comprising states $\{s_t\}$, observations $\{o_t\}$, and actions $\{\alpha_t\}$. At any given timestep $t$, the policy $\pi$ receives the current observation $o_t$ and predicts a sequence of future actions $\alpha_{t:t+T-1}$. We refer to this fixed-length sequence of size $T$ as an \textit{action chunk}.}

As shown in Figure~\ref{fig:action-rate-illustration}, we demonstrate the \change{action rate} multimodal distribution through an example of a sandwich making task. While two expert demonstrations follow identical spatial trajectories from point A to point D, their varying execution speeds result in different action chunk representations: [A→B] versus [A→B→C→D], causing spatially equivalent motions to be treated as distinct training samples.

To address \change{action rate} bias, we initially consider two straightforward approaches. The first method normalizes all demonstrations to uniform \change{action rate}, ensuring consistent spatial distance per action chunk. However, this approach fails to capture task-specific \change{action rate} requirements: fine-grained tasks like plug insertion require precise alignment phases, while pouring tasks necessitate deliberate pauses during execution. The second approach rescales all demonstrations of a task to identical temporal duration~\cite{masuya2025variable}, but this episode-level temporal normalization cannot eliminate \change{action rate} distribution bias due to heterogeneous speed patterns across trajectory segments within episodes and the inherent requirement for varying episode lengths under different initial conditions.

\change{Action rate} bias destabilizes training: identical inputs are mapped to action chunks with inconsistent \change{action rate} distributions across demonstrations. To address this issue, we propose an \change{Action Rate Model (ARM)} that predicts the expected robot \change{action rate} conditioned on observations $\mathbf{o}_t$. We train the \change{ARM} using MSE loss:
\begin{equation}
\mathcal{L}_{\text{ARM}} = \mathbb{E}_{(\mathbf{o}_t, \mathbf{a}_{t:t+T}) \sim \mathcal{D}} \left[ \|\text{ARM}(\mathbf{o_t}) - v(\mathbf{a_{t:t+T}})\|_2^2 \right],
\end{equation}
where $v(\mathbf{a_{t:t+T}})$ denotes the \change{action rate} metric extracted from action sequence $\mathbf{a_{t:t+T}}$, and $\mathcal{D}$ represents the demonstration dataset. This formulation enables the \change{ARM} to learn the expected \change{action rate} profile for demonstrations with similar observations.

Specifically, we define $v(\mathbf{a_{t:t+T}})$ based on the end-effector relative displacement representation. Let $\mathbf{a}^{\text{eef}}_{t:t+T} \in \mathbb{R}^{T \times D}$ denote the end-effector actions, where $T$ represents the action chunk size and $D$ corresponds to the degrees of freedom. We initially normalize each dimension of $\mathbf{a}^{\text{eef}}$ to the range $[-1, 1]$, subsequently defining the \change{action rate} metric as:
\begin{equation}
v(\mathbf{a_{t:t+T}}) = \|\mathbf{a}^{\text{eef}}_{t:t+T}\|_1,
\end{equation}
where $\|\cdot\|_1$ denotes the L1 norm.

Figure~\ref{fig:go-1-pro} illustrates the complete process of employing our \change{ARM} for distribution debiasing. During policy training via imitation learning, for each training sample $(\mathbf{o}_t, \mathbf{a}_{t:t+T})$, we determine the optimal chunk length $L$ by: 
\begin{equation}
L = \arg\min_L |\text{ARM}(\mathbf{o}_t) - v(\mathbf{a}_{t:t+L})|.
\end{equation}
To ensure training stability and mitigate potential interference from \change{ARM} prediction errors, we constrain the search range of $L$ to lie within $0.5T$ and $1.5T$. Subsequently, we employ interpolation to transform $\mathbf{a}_{t:t+L}$ into $\tilde{\mathbf{a}}_{t:t+T}$ with the desired chunk size $T$. This temporal rescaling ensures that all training samples with similar observations $\mathbf{o}$ exhibit consistent \change{action rates}, thereby achieving \change{action rate} distribution debiasing.

Our \change{ARM} employs a simple yet effective architecture, consisting of a SigLIP~\cite{zhai2023siglip} visual encoder followed by an MLP head. The \change{ARM} processes three input images through SigLIP to extract visual features, which are subsequently mapped to a scalar \change{action rate} value via the MLP. We normalize the output \change{action rate} to [0,1] using min-max scaling to enhance training stability. Critically, we freeze the SigLIP encoder during training to prevent the \change{ARM} from overfitting to fine-grained visual details in training samples, thereby ensuring it learn to predict the average \change{action rate} for similar observations rather than memorizing specific visual patterns. 

\change{Importantly, implementing action rate distribution debiasing imposes specific requirements on the underlying robot controller. Direct interpolation of action rates necessitates a high-precision controller to ensure that the modified dynamics are tracked accurately, as highlighted in concurrent works focusing on demonstration acceleration~\cite{guo2025demospeedup, arachchige2025sail}. To mitigate this in scenarios with lower-precision hardware, we can adopt the strategy proposed by Arachchige et al.~\cite{arachchige2025sail}: training the policy to predict \textit{reached poses} (actual robot states) rather than \textit{commanded poses}, effectively decoupling policy learning from the imperfections of the data collection controller, allowing a high-fidelity tracking controller to execute the predicted trajectory during deployment.}

\begin{table*}[t]
\centering
\caption{\change{\textbf{Performance evaluation using 10\% episode-based sampling from AgiBot World Beta dataset for pre-training.} Distribution debiasing applied during the pre-training phase demonstrates consistent performance improvements. Additional debiasing during fine-tuning further enhances model capabilities across all evaluated tasks. The improvements from distribution debiasing show statistically significant differences (\(p < 0.05\)).}}
\begin{tabular}{cc!{\vrule}cccc!{\vrule}c}
\toprule[1pt]
\multicolumn{2}{c!{\vrule}}{Training Setting} & \multicolumn{4}{c!{\vrule}}{Task Completion Score} & Average \\
\midrule
Pre-training Data     & Fine-tuning Data     & Pour Water & Fold Shorts & Make Sandwich & Wipe Table & Average \\
\midrule
Biased            & Biased            & 0.20\change{$\pm$0.02}       & 0.30\change{$\pm$0.03}        & 0.67\change{$\pm$0.04}          & 0.66\change{$\pm$0.04}       & 0.46\change{$\pm$0.04}    \\
Debiased          & Biased            & 0.27\change{$\pm$0.02}       & 0.30\change{$\pm$0.02}        & 0.71\change{$\pm$0.03}          & 0.68\change{$\pm$0.04}       & 0.49\change{$\pm$0.03}    \\
Biased          & Debiased            & 0.26\change{$\pm$0.02}       & 0.33\change{$\pm$0.03}        & \textbf{0.73\change{$\pm$0.02}}          & 0.69\change{$\pm$0.04}       & 0.50\change{$\pm$0.03}    \\
Debiased          & Debiased          & \textbf{0.32\change{$\pm$0.03}}       & \textbf{0.37\change{$\pm$0.04}}        & \textbf{0.73\change{$\pm$0.04}}          & \textbf{0.70\change{$\pm$0.03}}       & \textbf{0.53\change{$\pm$0.03}}    \\
\bottomrule[1pt]
\end{tabular}
\label{tab:distribution-debias}
\end{table*}

\subsection{Distribution Debiasing in the Pre-training Phase}

We first investigate how distribution debiasing influences model performance across different training stages using 10\% episode-based sampling from the AgiBot World Beta dataset.

As presented in Table~\ref{tab:distribution-debias}, applying distribution debiasing exclusively during pre-training yields a 6.5\% average improvement (from 0.46 to 0.49), with particularly notable gains in Pour Water (+35\%). However, this configuration introduces a distribution mismatch: the debiased pre-trained representations must adapt to biased fine-tuning data, which constrains potential performance gains and may introduce training instability.

More substantial improvements are observed when distribution debiasing is consistently applied across both pre-training and fine-tuning stages. This unified debiasing strategy achieves a 15\% overall improvement (from 0.46 to 0.53), with Pour Water showing the most significant enhancement (+60\% from 0.20 to 0.32). Notably, this performance gain equals that achieved by scaling the pre-training dataset by 2.5× (as demonstrated in Figure~\ref{fig:pre-train-his}), underscoring the data efficiency benefits of our debiasing method.

The task-specific improvements reveal distinct patterns: manipulation-heavy tasks such as Pour Water and Fold Shorts benefit more from consistent debiasing (+60\% and +23\% respectively), while navigation-oriented tasks show more modest but meaningful gains. This suggests that distribution debiasing is particularly effective for tasks requiring precise action sequences and complex manipulation strategies, where biased demonstrations can impede learning efficiency.

\subsection{Distribution Debiasing in the Fine-tuning Phase}

\begin{figure}[t]
\centering
\includegraphics[width=\columnwidth]{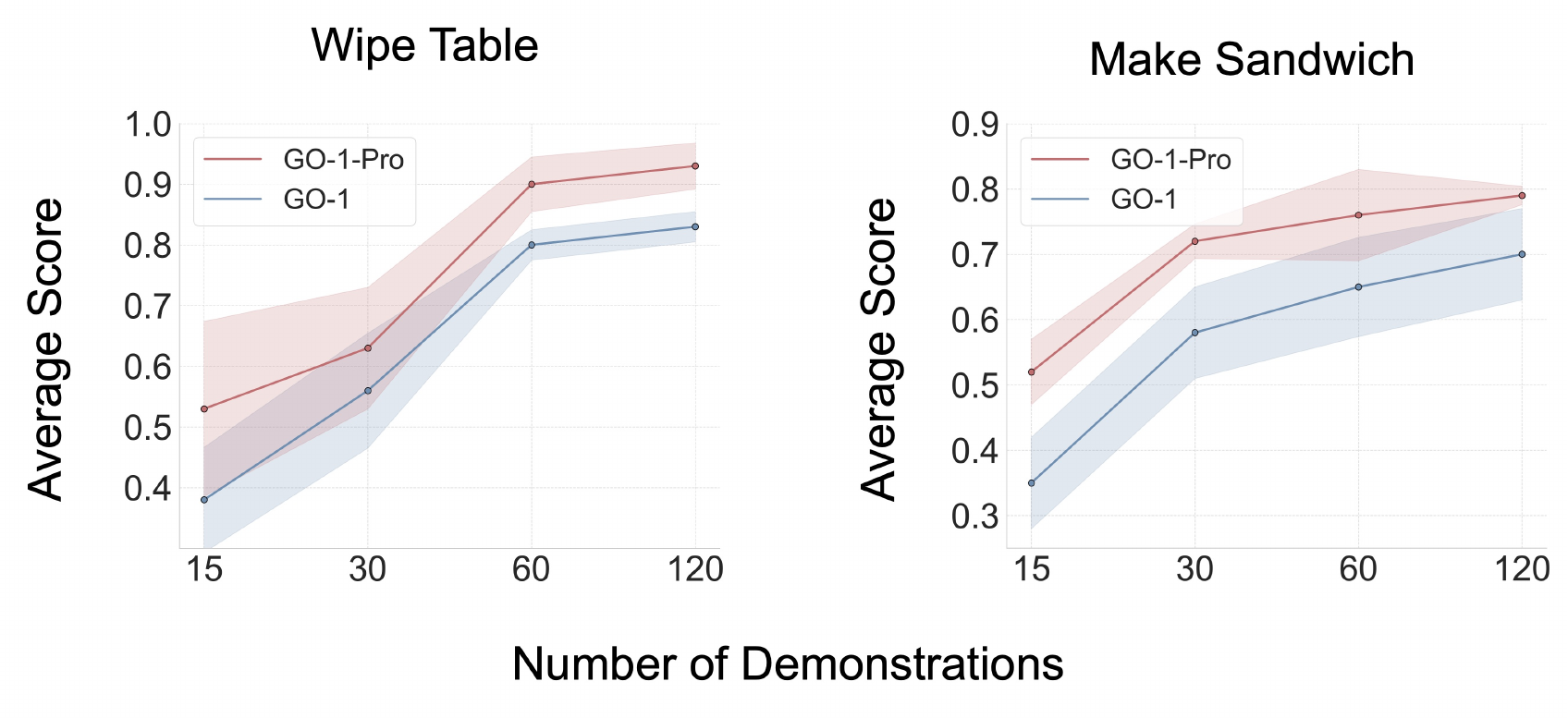}
\caption{\textbf{GO-1-Pro consistently outperforms GO-1 on both the Wipe Table and Make Sandwich tasks.} GO-1-Pro achieves comparable results using only 50\% of the training data that GO-1 uses, demonstrating superior data efficiency.}
\label{fig:dd-sft-score}
\end{figure}

\begin{figure}[t]
\centering
\includegraphics[width=\columnwidth]{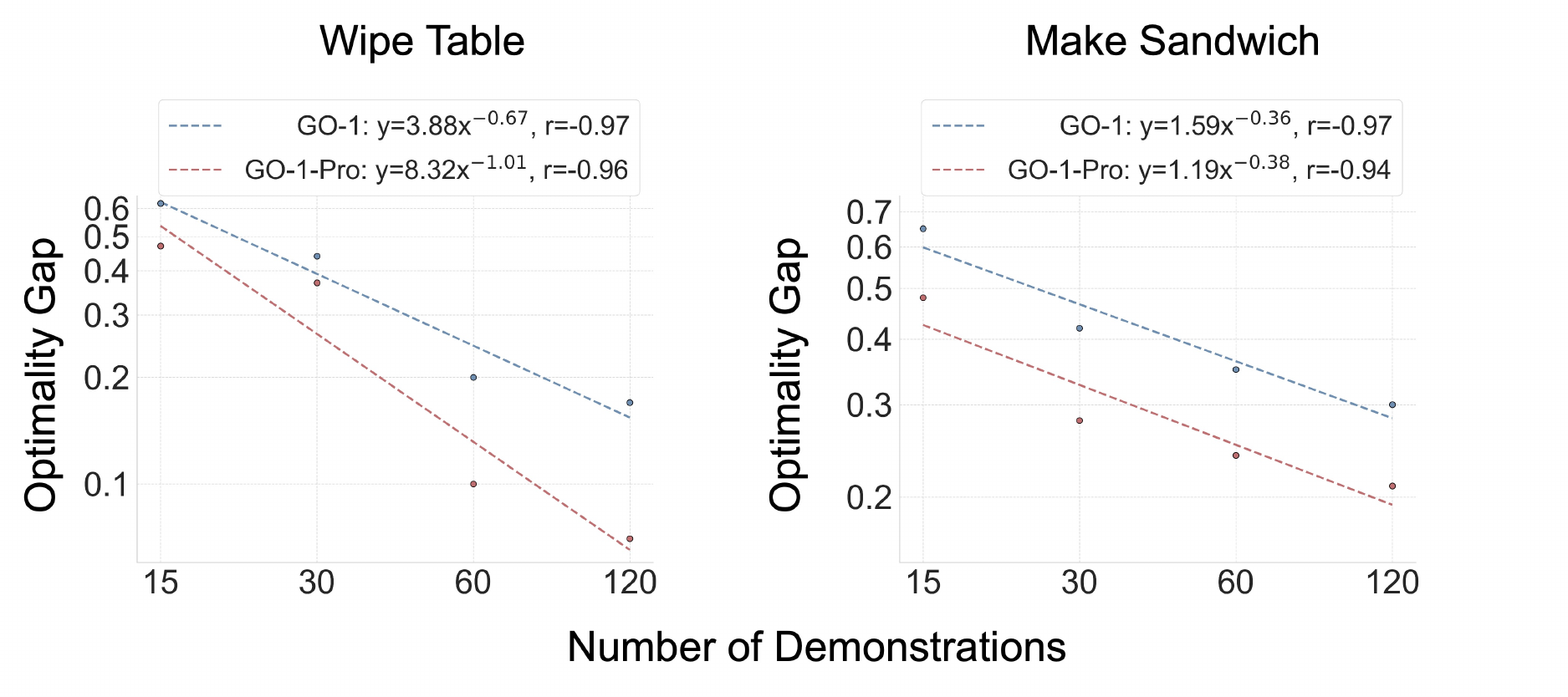}
\caption{\textbf{Performance scaling with fine-tuning data usage for GO-1 and GO-1-Pro on Wipe Table and Make Sandwich tasks.} Both models follow power law relationships (fitted equations shown), with GO-1-Pro demonstrating faster convergence rates}
\label{fig:dd-sft-power-law}
\end{figure}

Due to the substantial computational cost of pre-training, in this section we explore the effectiveness of applying distribution debiasing exclusively during the fine-tuning stage. For the Wipe Table and Make Sandwich tasks, we evaluate both tasks using the GO-1 model pre-trained on the full AgiBot World Beta dataset under different fine-tuning data scales. We refer to the model fine-tuned on distribution-debiased data as GO-1-Pro. The experimental results presented in Figure~\ref{fig:dd-sft-score} demonstrate that GO-1-Pro consistently outperforms GO-1 across both tasks and all data scales, achieving an average score of 0.93 on Wipe Table compared to GO-1's 0.83, and reaching 0.79 on Make Sandwich while GO-1 plateaus at 0.7.

Notably, GO-1-Pro exhibits exceptional data efficiency—it achieves comparable or superior performance using only half the training data required by GO-1. Specifically, GO-1-Pro with 60 demonstrations outperforms GO-1 with 120 demonstrations on both tasks, effectively doubling data utilization efficiency. These results underscore the critical importance of addressing data distribution biases in robotic learning.

The benefits of our distribution debiasing approach become particularly pronounced in low-data regimes, where GO-1-Pro improves performance from 0.35 to 0.52 for Make Sandwich and from 0.38 to 0.53 for Wipe Table with only 15 demonstrations. Under data-scarce conditions, the multimodal distribution across \change{action rate} and spatial dimensions creates substantial interference in the model's learning process, impeding its ability to effectively capture essential spatial distribution patterns. By disentangling these confounding factors, our distribution debiasing method enables the model to focus on learning core spatial relationships despite limited data availability, thereby facilitating more efficient and robust policy learning.

To investigate how distribution debiasing methods affect model performance across different fine-tuning data scales, we fit power-law curves to analyze the impact of fine-tuning data scale on final model performance, with results presented in Figure~\ref{fig:dd-sft-power-law}. Both GO-1 and GO-1-Pro exhibit power-law scaling behavior across the two tasks, but with notably different characteristics. For the Wipe Table task, GO-1-Pro demonstrates significantly faster convergence with an exponent of -1.01 compared to GO-1's -0.67, indicating that GO-1-Pro achieves near-optimal performance more rapidly as data volume increases. The steeper negative exponent suggests that our distribution debiasing method more effectively leverages additional training data to reduce the optimality gap. For the Make Sandwich task, while both models exhibit similar exponents (-0.38 for GO-1-Pro vs -0.36 for GO-1), GO-1-Pro maintains consistently lower optimality gaps across all data scales. This parallel scaling with a constant performance offset indicates that the benefits of distribution debiasing persist regardless of data volume.

%% file: text_annotated/conclusion.tex
This work systematically investigates data scaling principles for robotic manipulation, revealing three key insights that challenge conventional wisdom. We find that (1) task diversity proves more critical than per-task demonstration quantity for effective transfer, (2) embodiment diversity is not mandatory for achieving cross-embodiment transfer capabilities, \minorrev{but data quality and consistency is essential},
and (3) expert diversity can be confounding due to \change{action rate} multimodality, leading us to propose a distribution debiasing method that yields substantial performance gains. These findings challenge the ``more diverse is better'' paradigm and provide practical guidance for strategically scaling robotic manipulation datasets.

\textbf{Limitations and future work.}  
While our distribution debiasing method successfully eliminates the \change{action rate} multimodality, it cannot be applied to dynamic tasks such as ping-pong where the varying \change{action rates} are crucial for robot-environment interaction. Additionally, there remain other aspects of expert diversity that harm policy learning, such as meaningless pauses during data collection and suboptimal behavioral patterns that could cause robots to enter infinite loops. Future work could further explore these areas by developing methods to identify and mitigate confounding expert diversity while preserving beneficial variations.

%% file: bio.tex
\begin{IEEEbiography}
[{\includegraphics[width=1in,height=1.25in,clip,keepaspectratio]{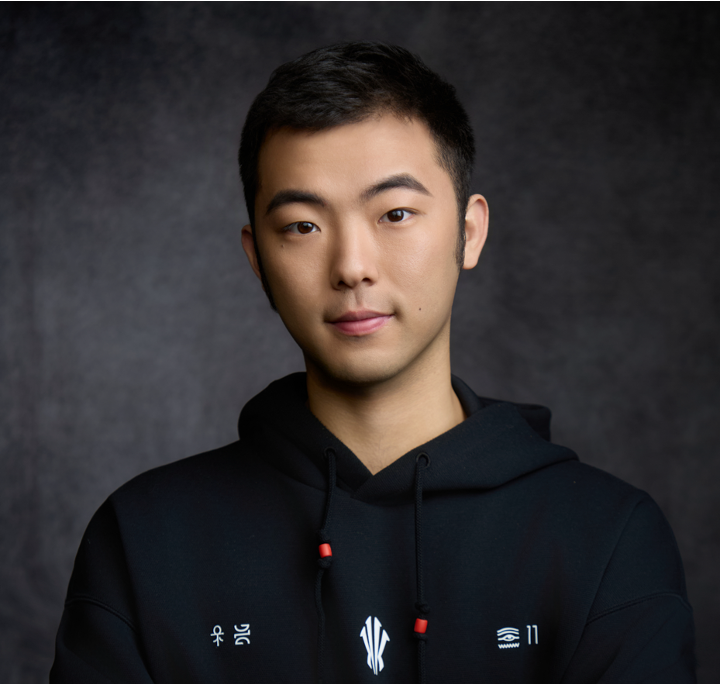}}]
{Modi Shi} received the B.S. degree from the School of Computer Science and Engineering, Beihang University, Beijing, China. He is currently pursuing the Ph.D. degree at the School of Computer Science and Engineering, Beihang University. His research interests include humanoids and robotic manipulation.
\end{IEEEbiography}

\begin{IEEEbiography}
[{\includegraphics[width=1in,height=1.25in,clip,keepaspectratio]{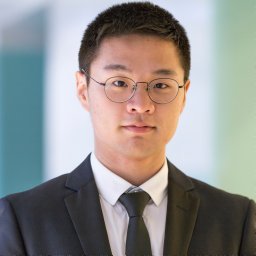}}]
{Li Chen} is currently a Ph.D. student at Department of Computer Science, the University of Hong Kong. He received the B.E. in mechanical engineering from Shanghai Jiao Tong University, and the M.S. in Robotics from the University of Michigan, Ann Arbor, USA. His research interests lie in physical AI including robotics and autonomous driving.
\end{IEEEbiography}

\begin{IEEEbiography}
[{\includegraphics[width=1in,height=1.25in,clip,keepaspectratio]{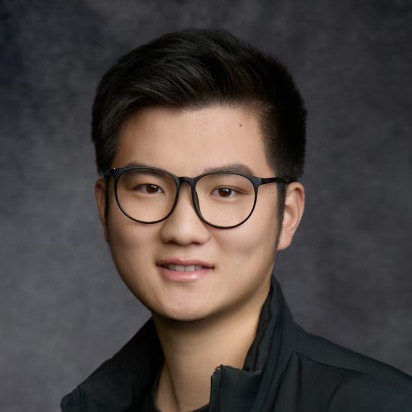}}]
{Jin Chen} is currently a Ph.D. student at Shanghai Innovation Institute and in Computer Engineering at Fudan University. He received the M.S. in Mathematics from Xi’an Jiaotong University, and the B.S. in Mathematics from China University of Mining and Technology. His research interests lie in embodied AI, specifically whole-body loco-manipulation.
\end{IEEEbiography}

\begin{IEEEbiography}
[{\includegraphics[width=1in,height=1.25in,clip,keepaspectratio]{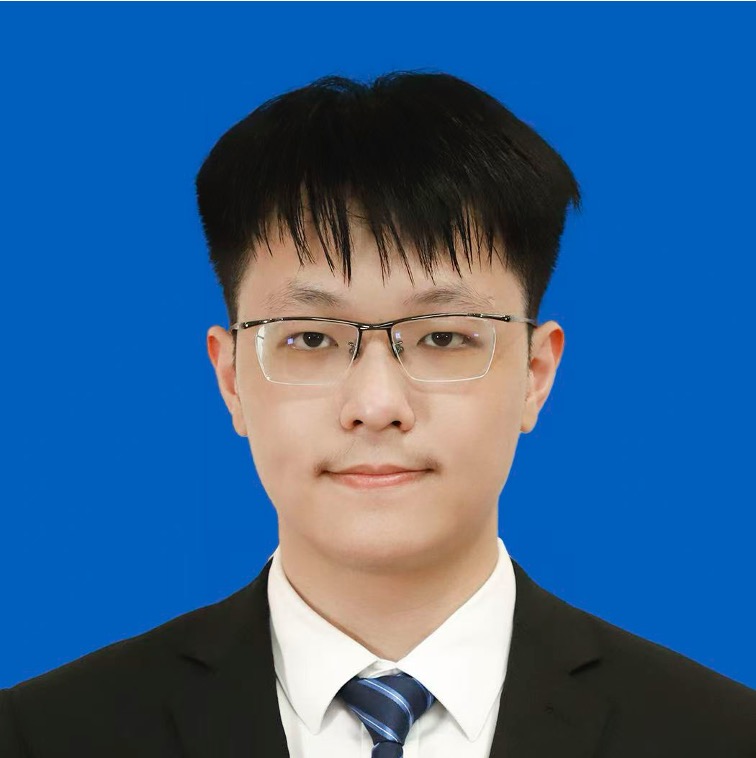}}]
{Yuxiang Lu} is currently a Ph.D. student at the School of Computing and Data Science, the University of Hong Kong. He received the M.E. and B.E. from Shanghai Jiao Tong University. His research interests include embodied AI and computer vision.
\end{IEEEbiography}

\begin{IEEEbiography}
[{\includegraphics[width=1in,height=1.25in,clip,keepaspectratio]{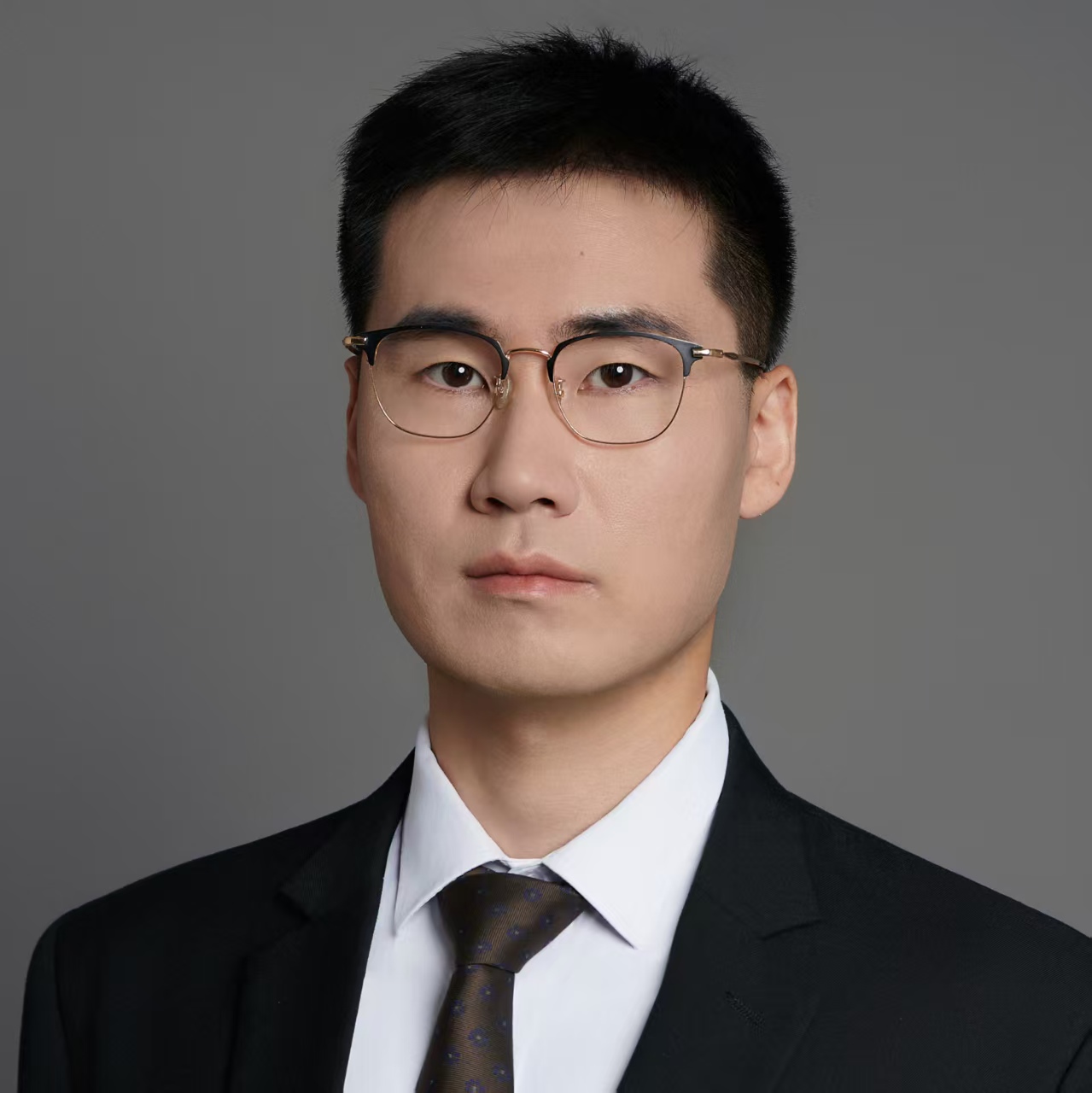}}]
{Chiming Liu} is a Large Model Architect at AgiBot. He was previously a Large Language Model (LLM) Architect at Biomap, and prior to that, an Expert Researcher at Tencent. He holds degrees from Nanjing University, China, and the University of Southampton, UK. His research interests lie in the design, pre-training, and post-training of large models, and in the development of large-scale embodied AI systems.
\end{IEEEbiography}

\begin{IEEEbiography}
[{\includegraphics[width=1in,height=1.25in,clip,keepaspectratio]{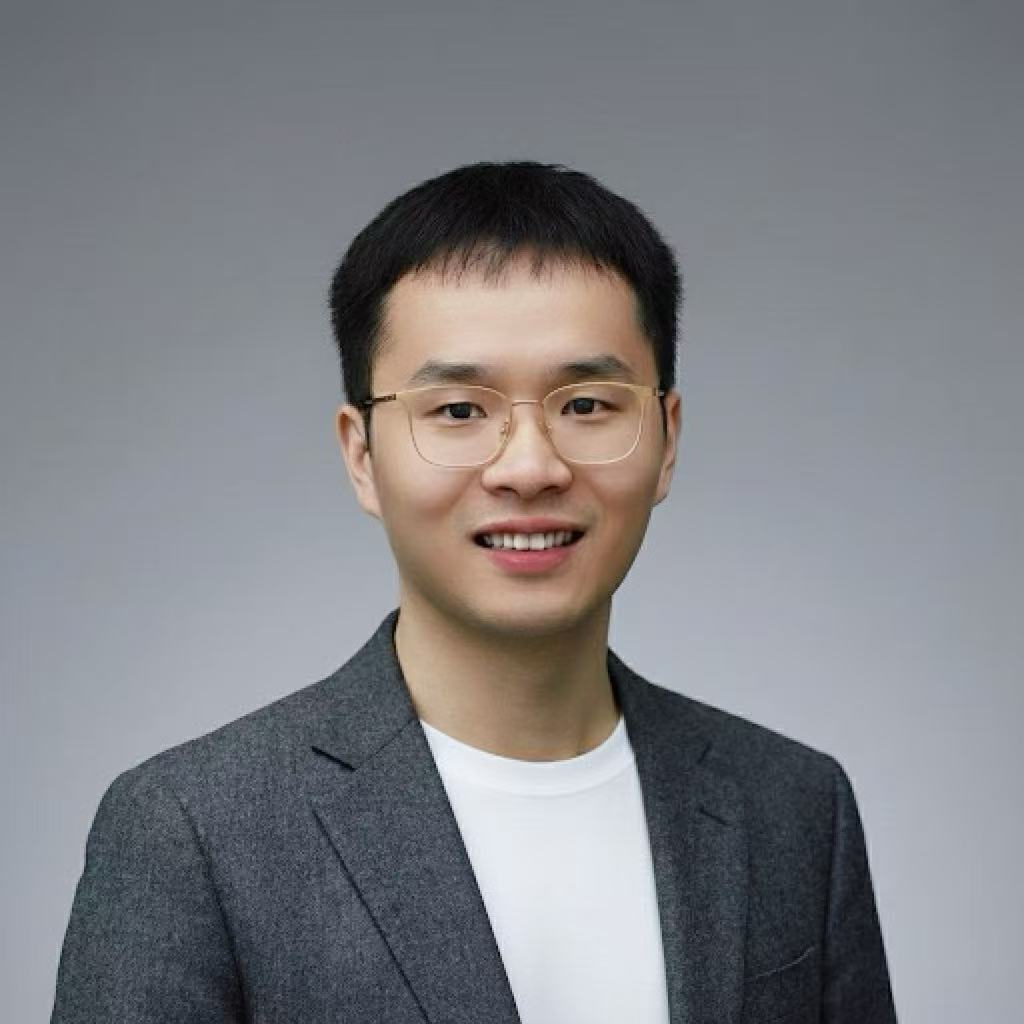}}]
{Guanghui Ren} is currently the Director of AI Algorithms at the Embodied Intelligence Business Unit of AgiBot. He received his master’s degree from the Chinese Academy of Sciences. His research focuses on multimodal large language models, embodied manipulation models, and embodied world models, aiming to develop general-purpose robots.
\end{IEEEbiography}

\begin{IEEEbiography}
[{\includegraphics[width=1in,height=1.25in,clip,keepaspectratio]{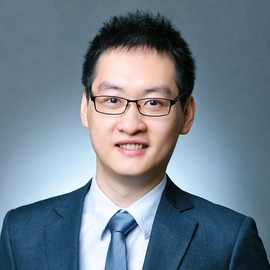}}]
{Ping Luo} received the PhD degree in information engineering, from the Chinese University of Hong Kong (CUHK), in 2014. He is an associate professor with the Department of Computer Science, University of Hong Kong (HKU). He was a postdoctoral fellow with CUHK from 2014 to 2016. He joined SenseTime Research as a Principal Research Scientist from 2017 to 2018. His research interests are machine learning and computer vision.
\end{IEEEbiography}

\begin{IEEEbiography}
[{\includegraphics[width=1in,height=1.25in,clip,keepaspectratio]{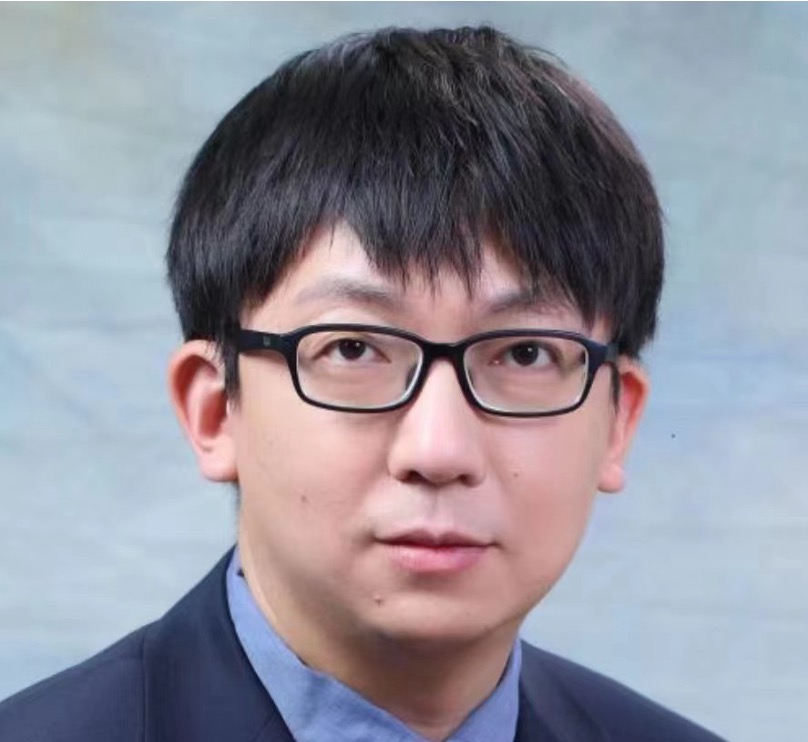}}] 
{Di Huang} received the B.S. and M.S. degrees in computer science from Beihang University, Beijing, China, in 2005 and 2008, respectively, and the Ph.D. degree in computer science from the \'{E}cole Centrale de Lyon, Lyon, France, in 2011. He joined School of Computer Science and Engineering, Beihang University, where he is currently a Professor. His research interests include biometrics, 2D/3D face analysis, image/video processing, and pattern recognition.
\end{IEEEbiography}

\begin{IEEEbiography}
[{\includegraphics[width=1in,height=1.25in,clip,keepaspectratio]{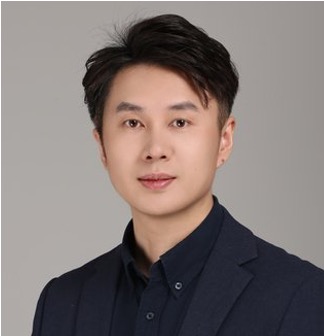}}] 
{Maoqing Yao} currently serves as Partner, Senior Vice President, and President of the Embodied Intelligence Business Unit at AgiBot. He received his bachelor’s degree from Tsinghua University and his Ph.D. from the University of Southern California. Previously, he held technical leadership positions at NIO, Waymo, Google, and Oracle. His research focuses on embodied intelligence, with expertise in AI and robotics research, development, and deployment.
\end{IEEEbiography}

\begin{IEEEbiography}
[{\includegraphics[width=1in,height=1.25in,clip,keepaspectratio]{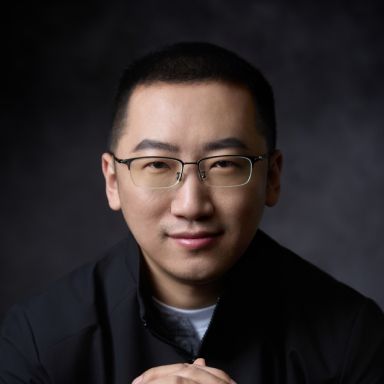}}]
{Hongyang Li} is an Assistant Professor at School of Computing and Data Science, University of Hong Kong. His research focuses on autonomous driving and embodied AI. He led the UniAD project, winning the IEEE CVPR 2023 Best Paper Award, and created AgiBot World for investigating scaling laws in robotic manipulation. He proposed BEVFormer, selected as Top 100 AI Papers in 2022. He served as Area Chair for CVPR, NeurIPS, ICLR, ICCV, ICML, and RSS, and is a Senior Member of IEEE. He received the China AI Wu Wen Jun Early Career Award 2024.
\end{IEEEbiography}

%% file: text_annotated/appendix.tex
\subsection{Hardware Setup}
\label{appendix:hardware-setup}

\begin{figure}[htbp]
\centering
\includegraphics[width=0.9\columnwidth]{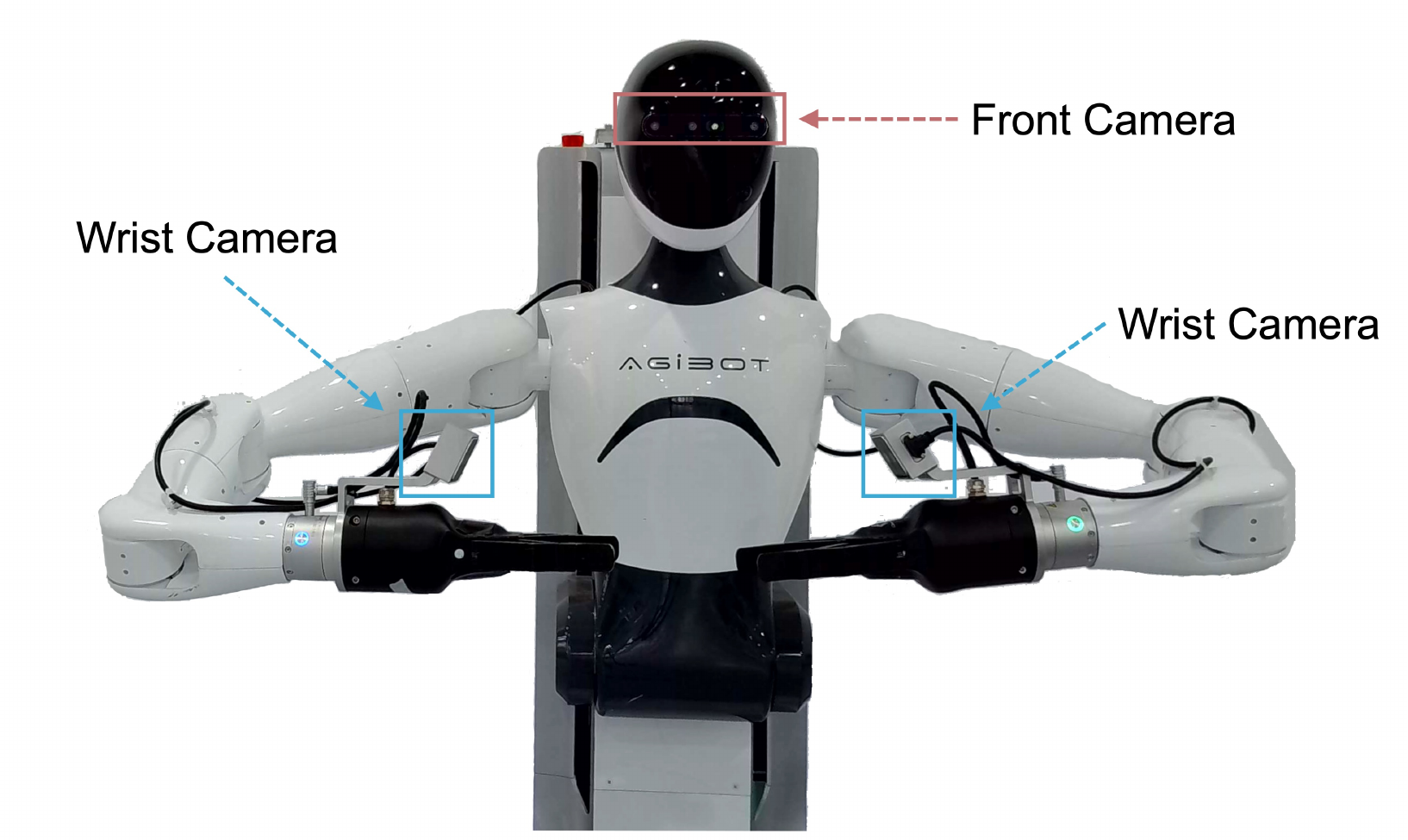}
\caption{Deployment on AgiBot G1.}
\label{fig:agibot_g1}
\end{figure}

\begin{figure}[htbp]
\centering
\includegraphics[width=0.9\columnwidth]{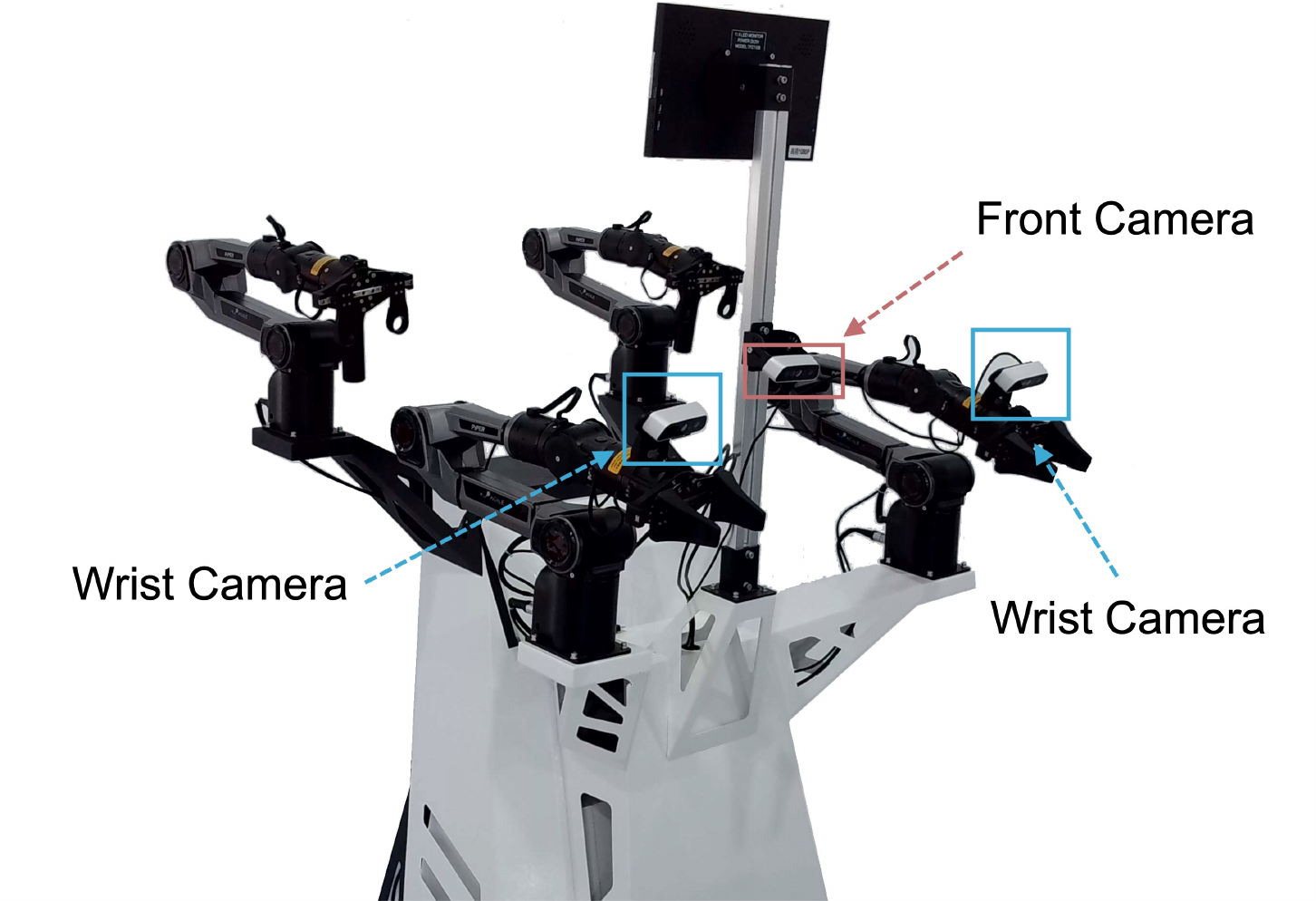}
\caption{Deployment on AgileX Cobot Magic with Piper.}
\label{fig:cobot_magic}
\end{figure}

Our real-world experiments are conducted on two robotic platforms: AgiBot G1 and AgileX Cobot Magic with Piper, as shown in Figure~\ref{fig:agibot_g1} and Figure~\ref{fig:cobot_magic} respectively. Both platforms are equipped with a multi-camera setup consisting of one front-facing camera mounted on the robot's head and two wrist cameras attached to each arm. 

\subsection{Implementation Details}
\subsubsection{Task Diversity}
\label{appendix:implentation-details-task-diversity}
Our experimental configurations vary according to the scale of pre-training data. For pre-training with 10\% of the AgiBot World dataset, we employ 32 H100 GPUs with a batch size of 64 per GPU, training for 100K steps. When scaling to 25\% of the data, we maintain the same hardware setup (32 GPUs, batch size 64 per GPU) but extend training to 250K steps. For full-scale pre-training on the complete dataset, we scale up to 96 H100 GPUs with batch size 64 per GPU, training for 500K steps. During the pre-training phase, we train the Vision-Language Model (VLM) component alongside other model parameters. For fine-tuning, we use 8 GPUs with a batch size of 128 per GPU, training for 20K steps while keeping the VLM parameters frozen.

\subsubsection{Embodiment Diversity}
Embodiment Diversity Experiments: We fine-tuned RDT-OXE and RDT-AWB on three platforms: Maniskill, RoboTwin, and real-world Agilex, with results reported in Figure~\ref{fig:mani-size-scale}, Figure~\ref{fig:twin-size-scale}, and Table~\ref{tab:agilex} respectively.

For Maniskill, we evaluated five tasks (PegInsertionSide, PickCube, StackCube, PlugCharger, PushCube) using only the third-person observations. We fine-tuned with 125/250/500/1000 demonstrations per task for 37.5K/75K/150K/300K steps respectively, saving checkpoints every 10K steps and reporting peak success rates.

For RoboTwin, we tested four tasks (BlockHammerBeat, BlocksStack, ContainerPlace, DualBottlesPick) using four camera views (front, head, left/right wrist). Fine-tuning used 25/50/100/200 demonstrations per task for 10K/30K/50K/60K steps respectively.

For real-world Agilex, we evaluated four tasks (PackageProduct, FoldShorts, CleanTrash, IndustrialSorting) using three camera views (front, left/right wrist). We used 100 demonstrations per task (200 for FoldShorts due to complexity) and trained for 100K steps.

Pre-training Setup: RDT-AWB was pre-trained on AgiBot World using 96 H100 GPUs (batch size 32 per GPU, 200K steps). RDT-OXE used the official checkpoint (48 H100 GPUs, batch size 32 per GPU, 1M steps). All other hyperparameters remained consistent between models.

\subsubsection{Expert Diversity}
Our \change{action rate} model employs a SigLIP vision encoder coupled with a three-layer MLP featuring LayerNorm, GELU activations, and 0.1 dropout. For fine-tuning experiments, we train the \change{action rate} model using 8 H100 GPUs with a batch size of 128 per GPU for 20K steps, while pre-training experiments employ the same hardware configuration but extend training to 100K steps. The GO-1-Pro training setup remains identical to GO-1.

\subsection{Additional Experiment Results}

\subsubsection{\change{Task Diversity}}
\label{appendix:additional-results-task-diversity}

\change{To provide a more granular analysis of how different pre-training data compositions affect policy performance across task execution, we present per-step scores in Table~\ref{tab:task_diversity_per_step_score}. Each task is decomposed into sequential manipulation steps (e.g., grasp, place, wipe), allowing evaluation of performance distribution throughout the entire task sequence rather than relying solely on average scores.}

\begin{table*}[htbp]
\centering
\caption{\change{Performance comparison across different pretraining data.}}
\label{tab:task_diversity_per_step_score}
\resizebox{\textwidth}{!}{
\fontsize{4}{5}\selectfont
\setlength{\tabcolsep}{4pt}
\begin{tabular}{l!{\vrule}cccccccc!{\vrule}cc}
\toprule[0.5pt]
\multirow{2}{*}{\textbf{Pretrain Data}} & \multicolumn{8}{c|}{\textbf{Make Sandwich}} & \multicolumn{2}{c}{\textbf{Wipe Table}} \\[-1.5pt]
\cmidrule(lr){2-9} \cmidrule(lr){10-11}
& Grasp & Place & Grasp & Place & Grasp & Place & Grasp & Place & Grasp & Wipe \\[-1pt]
\midrule
No Pretrain & 0.53 & 0.54 & 0.43 & 0.45 & 0.36 & 0.30 & 0.17 & 0.12 & 0.75 & 0.41 \\
10\% Scenario & 0.63 & 0.60 & 0.37 & 0.37 & 0.31 & 0.27 & 0.23 & 0.18 & 0.84 & 0.48 \\
10\% Task & 0.70 & 0.60 & 0.43 & 0.40 & 0.35 & 0.33 & 0.27 & 0.20 & 0.82 & 0.48 \\
10\% Episode & 0.80 & 0.83 & 0.76 & 0.73 & 0.63 & 0.67 & 0.46 & 0.48 & 0.82 & 0.50 \\
25\% Episode & 0.83 & 0.80 & 0.76 & 0.70 & 0.67 & 0.63 & 0.57 & 0.48 & 0.85 & 0.67 \\
Full Data & 0.86 & 0.80 & 0.83 & 0.73 & 0.63 & 0.60 & 0.57 & 0.57 & 0.92 & 0.73 \\
\midrule
\end{tabular}
}

\vspace{-0.5mm}

\resizebox{\textwidth}{!}{
\fontsize{4}{5}\selectfont
\setlength{\tabcolsep}{4pt}
\begin{tabular}{l|cccccc|ccc|c}
\multirow{2}{*}{\textbf{}} & \multicolumn{6}{c|}{\textbf{Fold Shorts}} & \multicolumn{3}{c|}{\textbf{Pour Water}} & \multirow{2}{*}{\textbf{Avg.}} \\[-1.5pt]
\cmidrule(lr){2-7} \cmidrule(lr){8-10}
& Grasp & Grasp & Fold & Grasp & Grasp & Fold & Grasp & Pour & Place & \\[-1pt]
\midrule
No Pretrain & 0.55 & 0.46 & 0.07 & 0.03 & 0.00 & 0.00 & 0.00 & 0.00 & 0.00 & 0.28 \\
10\% Scenario & 0.53 & 0.47 & 0.23 & 0.09 & 0.06 & 0.00 & 0.10 & 0.03 & 0.00 & 0.33 \\
10\% Task & 0.80 & 0.60 & 0.27 & 0.13 & 0.10 & 0.07 & 0.13 & 0.06 & 0.00 & 0.36 \\
10\% Episode & 0.73 & 0.60 & 0.23 & 0.13 & 0.07 & 0.03 & 0.30 & 0.17 & 0.13 & 0.46 \\
25\% Episode & 0.77 & 0.74 & 0.13 & 0.09 & 0.07 & 0.00 & 0.53 & 0.37 & 0.18 & 0.52 \\
Full Data & 0.80 & 0.77 & 0.43 & 0.33 & 0.27 & 0.22 & 0.50 & 0.33 & 0.19 & 0.58 \\
\bottomrule[0.5pt]
\end{tabular}
}
\end{table*}

\change{Furthermore, to verify that our findings on task diversity generalize beyond the GO-1 architecture, we conduct additional experiments using RDT~\cite{rdt} as an alternative policy model. We employ identical experimental settings as described in Section~\ref{sec:task-diversity}: the same three pre-training datasets (scenario-based, task-based, and episode-based sampling at 10\% scale), the same evaluation tasks (Wipe Table, Fold Shorts, Pour Water, Make Sandwich), and the same evaluation protocols across three scenarios (in-domain, visual distraction, and object-environment generalization).}

\begin{figure*}[htbp]
\centering
\includegraphics[width=\textwidth]{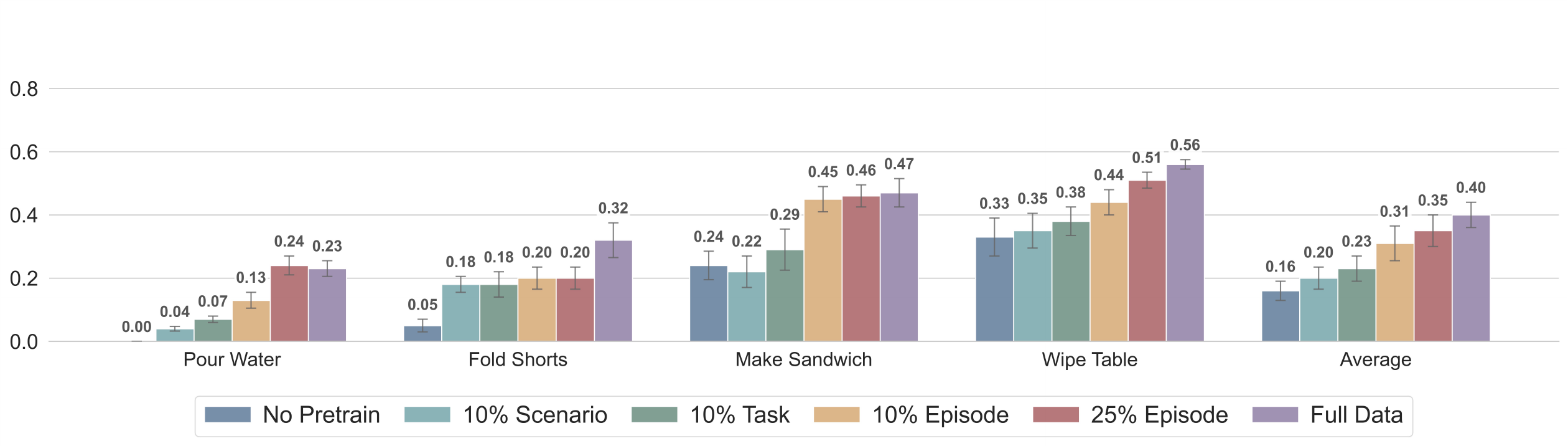}
\caption{\change{\textbf{Real-robot evaluation of RDT on four challenging tasks with different pre-training datasets.} Similar to GO-1 results, episode-based sampling consistently outperforms both task-based and scenario-based sampling, demonstrating that the importance of scene diversity generalizes across different model architectures.}}
\label{fig:rdt-pretrain-his}
\end{figure*}

\begin{figure}[htbp]
\centering
\includegraphics[width=\columnwidth]{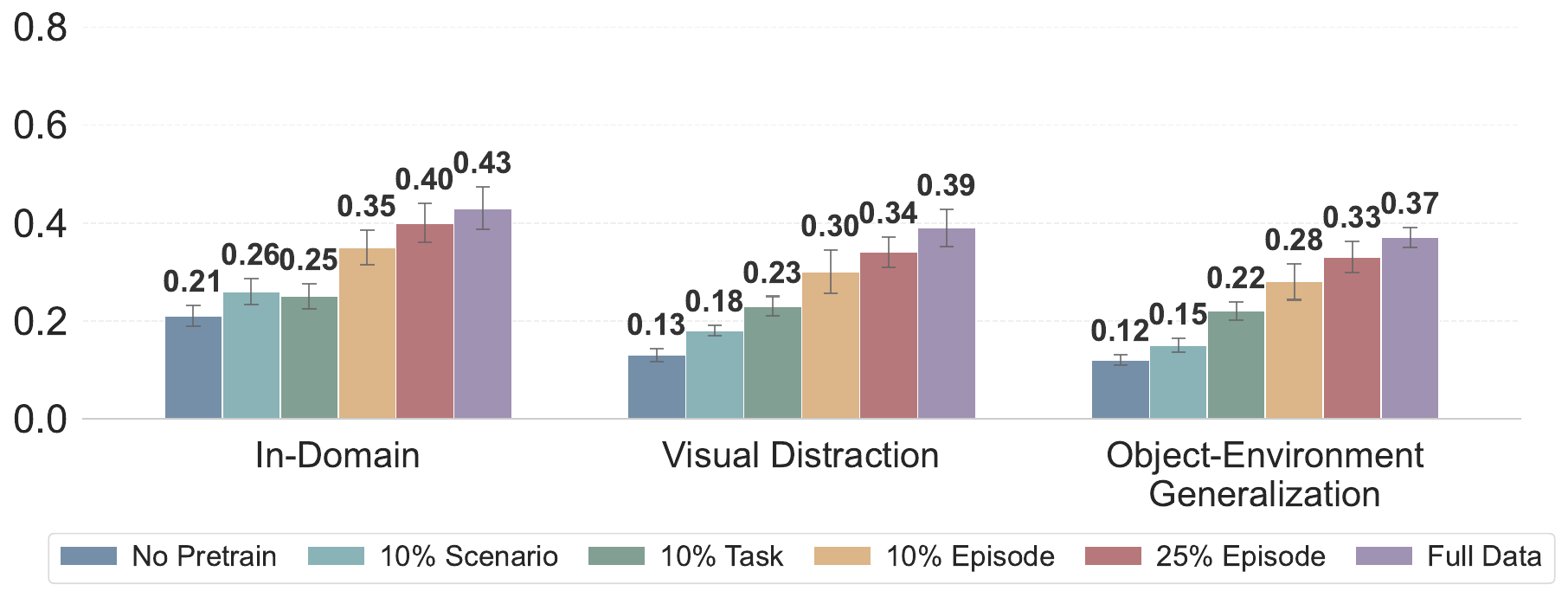}
\caption{\change{\textbf{RDT generalization performance across different pretraining strategies.} The model exhibits similar patterns to GO-1: episode-based sampling demonstrates superior performance under distribution shifts, with scene diversity providing greater robustness benefits than skill diversity alone.}}
\label{fig:rdt-gen-his}
\end{figure}

\change{As shown in Figure~\ref{fig:rdt-pretrain-his}, RDT exhibits consistent patterns with GO-1 findings. At 10\% data scale, scenario-based sampling achieves 0.20 average performance, task-based sampling improves to 0.23, while episode-based sampling reaches 0.31. Figure~\ref{fig:rdt-gen-his} further validates our conclusions about scene diversity versus skill diversity. For in-domain evaluation, episode-based sampling (0.35) outperforms task-based (0.25, +0.10) and scenario-based (0.26, +0.09) by similar margins. Under distribution shifts, scene diversity advantages become more pronounced: for visual distraction, episode-based (0.30) exceeds scenario-based (0.18) by 0.12 compared to 0.07 over task-based (0.23). For object-environment generalization, episode-based (0.28) surpasses scenario-based (0.15) by 0.13 versus 0.06 over task-based (0.22).}

\change{These results with RDT corroborate our main findings from GO-1 experiments: \textbf{(1)} scene diversity is more critical than skill diversity for robustness and generalization; \textbf{(2)} while both diversity types provide comparable in-domain benefits, scene diversity consistently provides greater advantages under distribution shifts; \textbf{(3)} these patterns generalize across different policy architectures, suggesting that the importance of diversity composition is a fundamental property of robotic manipulation learning rather than architecture-specific behavior.}

\subsubsection{Embodiment Diversity}
In this section, we provide additional detailed results for the simulation experiments in Section~\ref{sec:embodiment diversity}, showcasing the performance of RDT-OXE and RDT-AWB on each task in ManiSkill (Table~\ref{tab:mani-detail}) and RoboTwin (Table~\ref{tab:twin-detail}).

\begin{table*}[htbp]
\centering
\caption{\textbf{Performance on each task in ManiSkill.} The performance of RDT-OXE and RDT-AWB fine-tuned with various numbers of demonstrations for each task. Higher scores are \textbf{bolded} for emphasis.}
\begin{tabular}{c!{\vrule}c!{\vrule}ccccc!{\vrule}c}
\toprule[1pt]
Model    & Demonstrations & PegInsertionSide & PickCube & StackCube & PlugCharger & PushCube & Average Score \\ 
\midrule
\multirow{4}{*}{RDT-OXE} & 125            & 0.00             & 0.06      & 0.09       & 0.00        & 0.80      & 0.22    \\
                         & 250            & 0.00             & 0.25      & 0.34       & 0.00        & 0.97      & 0.31    \\
                         & 500            & 0.04             & 0.40      & 0.42       & 0.03        & 0.98      & 0.38    \\
                         & 1000           & 0.04             & 0.76      & 0.66       & 0.04        & 1.00      & 0.50    \\
\midrule[1pt]
\multirow{4}{*}{RDT-AWB} & 125            & 0.02             & 0.18      & 0.07       & 0.00        & 0.78      & 0.21    \\
                         & 250            & 0.02             & 0.19      & 0.40       & 0.00        & 0.96      & 0.32    \\
                         & 500            & 0.00             & 0.54      & 0.65       & 0.01        & 0.96      & 0.43    \\
                         & 1000           & 0.12             & 0.86      & 0.90       & 0.03        & 1.00      & 0.58    \\
\bottomrule[1pt]
\end{tabular}
\label{tab:mani-detail}
\end{table*}

\begin{table*}[htbp]
\centering
\caption{\textbf{Performance on each task in RoboTwin.} The performance of RDT-OXE and RDT-AWB fine-tuned with various numbers of demonstrations for each task. Higher scores are \textbf{bolded} for emphasis.}
\begin{tabular}{c!{\vrule}c!{\vrule}cccc!{\vrule}c}
\toprule[1pt]
Model   & Demonstrations & BlockHammerBeat & BlocksStack & ContainerPlace & DualBottlesPick & Average Score \\ 
\midrule
\multirow{4}{*}{RDT-OXE} & 25            & 0.55             & 0.02      & 0.52       & 0.19        & 0.32    \\
                         & 50            & 0.61             & 0.13      & 0.49       & 0.47        & 0.42    \\
                         & 100           & 0.87             & 0.32      & 0.57       & 0.58        & 0.59    \\
                         & 200           & 0.60             & 0.53      & 0.49       & 0.68        & 0.58    \\
\midrule[1pt]
\multirow{4}{*}{RDT-AWB} & 25            & 0.63             & 0.04      & 0.43       & 0.42        & 0.38    \\
                         & 50            & 0.67             & 0.20      & 0.40       & 0.42        & 0.42    \\
                         & 100           & 0.90             & 0.51      & 0.50       & 0.60        & 0.63    \\
                         & 200           & 0.94             & 0.55      & 0.60       & 0.67        & 0.69    \\
\bottomrule[1pt]
\end{tabular}
\label{tab:twin-detail}
\end{table*}

\subsubsection{Expert Diversity}

\change{To further validate the effectiveness of distribution debiasing at scale, we conduct additional experiments using the full AgiBot World Beta dataset for pre-training. As pre-training data scale increases, the action rate distribution becomes more complex and heterogeneous, potentially amplifying the benefits of distribution debiasing. As shown in Table~\ref{tab:distribution-debias-full}, distribution debiasing applied across both pre-training and fine-tuning stages yields the best performance (0.64 average score), consistent with our findings on 10\% data. Notably, the absolute performance improvements remain substantial even at this larger scale, demonstrating that debiasing benefits persist and potentially amplify as dataset complexity increases. This validates that our distribution debiasing approach is effective across different data scales and that addressing action rate multimodality becomes increasingly important as pre-training datasets grow in size and diversity.}
\begin{table*}[htbp]
\centering
\caption{\change{\textbf{Performance evaluation using full AgiBot World Beta dataset for pre-training.} Distribution debiasing applied during the pre-training phase demonstrates consistent performance improvements. Additional debiasing during fine-tuning further enhances model capabilities across all evaluated tasks.}}
\begin{tabular}{cc!{\vrule}cccc!{\vrule}c}
\toprule[1pt]
\multicolumn{2}{c!{\vrule}}{Training Setting} & \multicolumn{4}{c!{\vrule}}{Task Completion Score} & Average \\
\midrule
Pre-training Data     & Fine-tuning Data     & Pour Water & Fold Shorts & Make Sandwich & Wipe Table & Average \\
\midrule
Biased            & Biased            & 0.34$\pm$0.03       & 0.47$\pm$0.06        & 0.70$\pm$0.04          & 0.83$\pm$0.03       & 0.58$\pm$0.04    \\
Debiased          & Biased            & 0.38$\pm$0.05       & 0.52$\pm$0.05        & 0.70$\pm$0.03          & 0.85$\pm$0.03       & 0.61$\pm$0.04    \\
Biased          & Debiased            & 0.41$\pm$0.04       & 0.49$\pm$0.04        & 0.72$\pm$0.03          & 0.85$\pm$0.02       & 0.62$\pm$0.03    \\
Debiased          & Debiased          & \textbf{0.44$\pm$0.03}       & \textbf{0.53$\pm$0.05}        & \textbf{0.73$\pm$0.04}          & \textbf{0.86$\pm$0.02}       & \textbf{0.64$\pm$0.03}    \\
\bottomrule[1pt]
\end{tabular}
\label{tab:distribution-debias-full}
\end{table*}

\subsection{\change{Evaluation} Details}
\label{appendix:scoring}
In all the simulation experiments, we use the simple success rate as the evaluation metric. For real-world experiments, we define more fine-grained evaluation metrics to more precisely compare the capabilities of the models within a limited number of rollouts. \change{To ensure unbiased and consistent evaluation, we implemented rigorous protocols: dedicated personnel responsible for full-time model evaluation were not informed about which policy was being evaluated during trials to prevent potential bias. We have three independent scorers in total, with all trials from one specific task scored by a single scorer to ensure consistency within each task and eliminate inter-scorer variability. Importantly, these scorers are also the developers of the scoring rubric, ensuring deep understanding and consistent application of the evaluation criteria.} In Section~\ref{sec:task-diversity}, Section~\ref{sec:embodiment diversity} and Section~\ref{sec:expert diversity}, we cover seven tasks: Wipe Table, Fold Shorts, Pour Water, Make Sandwich, Package Product, Clean Trash, and Industrial Sorting. We have defined different evaluation metrics for each of them:

\subsubsection{Wipe Table}

\textbf{Task description.}
The robot is tasked with a contact-rich manipulation: Wipe Table, which demands the use of a sponge to clean beverage stains from the table surface. The task consists of 2 steps: first, the right arm grasps the sponge; and second, the right arm wipes the stain clean. The sponge, being a deformable object, introduces complexities in grasping and manipulation due to its softness and flexibility. The stain, being a liquid, presents additional challenges due to its irregular shape and the need for precise contact control.

\textbf{Scoring criteria.}
\begin{itemize}
\item \textbf{Step 1: Grasping the sponge}
\begin{itemize}
\item \textbf{scoring 0:} The gripper does not grasp the sponge.
\item \textbf{scoring 0.5:} The gripper attempts to grasp the sponge multiple times, struggles to maintain a stable grasp, but eventually succeeds.
\item \textbf{scoring 1:} The gripper successfully grasps the sponge without any slippage on the first attempt.
\end{itemize}
\item \textbf{Step 2: Wiping the stain clean}
\begin{itemize}
\item \textbf{scoring 0:} The sponge does not make contact with the stain, or the sponge drops during the wiping process.
\item \textbf{scoring 0.5:} The sponge makes contact with the stain but does not effectively wipe it clean, or the stain is only partially cleaned.
\item \textbf{scoring 1:} The sponge successfully wipes the stain completely clean on the first attempt.
\end{itemize}
\end{itemize}

\subsubsection{Fold Shorts}

\textbf{Task description.}
Fold Shorts is a complex, bimanual deformable object manipulation task that involves multiple folding operations to neatly organize shorts. This task requires the robot to perform a series of coordinated actions using both arms to fold the shorts accurately. The task consists of six steps: first, the left arm grasps the left side of the shorts; second, the right arm grasps the right side of the shorts; third, both arms fold the shorts forward; fourth, the left arm grasps the left side of the shorts again; fifth, the right arm grasps the right side of the shorts again; and sixth, both arms fold the shorts again, with the left arm holding the shorts in place and the right arm folding them over to the left arm's position. 
The complexity of this task lies in the need for precise bimanual coordination and the challenges posed by the deformable nature of the shorts. The softness and flexibility of the shorts make grasping and folding operations difficult, requiring careful control of force and motion. Additionally, the thin material of the shorts increases the risk of collisions with the table, adding another layer of complexity.

\textbf{Scoring criteria.}
\begin{itemize}
\item \textbf{Step 1: Left arm grasping the left side}
\begin{itemize}
\item \textbf{scoring 0:} The left arm does not grasp the shorts.
\item \textbf{scoring 0.5:} The left arm attempts to grasp the shorts multiple times, struggles to maintain a stable grasp, or takes too long to succeed.
\item \textbf{scoring 1:} The left arm successfully grasps the shorts on the first attempt without any slippage.
\end{itemize}
\item \textbf{Step 2: Right arm grasping the right side}
\begin{itemize}
\item \textbf{scoring 0:} The right arm does not grasp the shorts.
\item \textbf{scoring 0.5:} The right arm attempts to grasp the shorts multiple times, struggles to maintain a stable grasp, or takes too long to succeed.
\item \textbf{scoring 1:} The right arm successfully grasps the shorts on the first attempt without any slippage.
\end{itemize}
\item \textbf{Step 3: Bimanual forward folding of the shorts}
\begin{itemize}
\item \textbf{scoring 0:} The shorts are not folded forward or the folding is incomplete.
\item \textbf{scoring 0.5:} The shorts are folded forward but not perfectly aligned or the folding is not symmetrical.
\item \textbf{scoring 1:} The shorts are folded forward perfectly, with symmetrical alignment.
\end{itemize}
\item \textbf{Step 4: Left arm grasping the left side again}
\begin{itemize}
\item \textbf{scoring 0:} The left arm does not grasp the shorts.
\item \textbf{scoring 0.5:} The left arm attempts to grasp the shorts multiple times, struggles to maintain a stable grasp, or takes too long to succeed.
\item \textbf{scoring 1:} The left arm successfully grasps the shorts on the first attempt without any slippage.
\end{itemize}
\item \textbf{Step 5: Right arm grasping the right side again}
\begin{itemize}
\item \textbf{scoring 0:} The right arm does not grasp the shorts.
\item \textbf{scoring 0.5:} The right arm attempts to grasp the shorts multiple times, struggles to maintain a stable grasp, or takes too long to succeed.
\item \textbf{scoring 1:} The right arm successfully grasps the shorts on the first attempt without any slippage.
\end{itemize}
\item \textbf{Step 6: Bimanual folding of the shorts with the left arm holding and the right arm folding}
\begin{itemize}
\item \textbf{scoring 0:} The shorts are not folded or the folding is incomplete.
\item \textbf{scoring 0.5:} The shorts are folded but not perfectly aligned or the folding is not symmetrical.
\item \textbf{scoring 1:} The shorts are folded perfectly, with symmetrical alignment.
\end{itemize}
\end{itemize}

\subsubsection{Pour Water}

\textbf{Task description.}
Pour Water is a fine-grained manipulation task requiring the robot to grasp a kettle handle and pour water into a cup. The task consists of two steps: first, the robot must grasp the kettle handle; and second, the robot must pour water from the kettle into the cup. The task requires precise control over the kettle's position and pouring angle to ensure accurate pouring. The complexity of this task lies in the need for precise spatial and temporal control during the pouring process, including accurate positioning, controlled pouring, and managing the flow rate of the water.

\textbf{Scoring criteria.}
\begin{itemize}
\item \textbf{Step 1: Grasping the kettle handle}
\begin{itemize}
\item \textbf{scoring 0:} The gripper does not grasp the kettle handle.
\item \textbf{scoring 0.5:} The gripper attempts to grasp the kettle handle multiple times, struggles to maintain a stable grasp, or takes too long to succeed.
\item \textbf{scoring 1:} The gripper successfully grasps the kettle handle on the first attempt without any slippage.
\end{itemize}
\item \textbf{Step 2: Pouring water from the kettle into the cup}
\begin{itemize}
\item \textbf{scoring 0:} The robot does not pour the water, or the water is poured outside the cup.
\item \textbf{scoring 0.5:} The robot pours the water but spills some water outside the cup, or the pouring process takes too long or requires multiple attempts to succeed.
\item \textbf{scoring 1:} The robot successfully pours the water into the cup without any spillage on the first attempt.
\end{itemize}
\end{itemize}

\subsubsection{Make Sandwich}

\textbf{Task description.}
Make Sandwich is a long-horizon task that sequentially involves picking up bread, ham, and lettuce to assemble a sandwich in proper order. The task consists of eight steps: first, the robot must grasp the first slice of bread; second, place the bread on the plate; third, grasp a slice of ham; fourth, place the ham on the bread; fifth, grasp a piece of lettuce; sixth, place the lettuce on the ham; seventh, grasp the second slice of bread; and eighth, place the bread on the lettuce. The complexity of this task lies in the need for precise sequential manipulation and the long-horizon nature of the task, requiring the robot to maintain accuracy and stability over multiple steps.

\textbf{Scoring criteria.}
\begin{itemize}
\item \textbf{Step 1: Grasping the first slice of bread}
\begin{itemize}
\item \textbf{scoring 0:} The gripper does not grasp the bread.
\item \textbf{scoring 0.5:} The gripper attempts to grasp the bread multiple times, struggles to maintain a stable grasp, or takes too long to succeed.
\item \textbf{scoring 1:} The gripper successfully grasps the bread on the first attempt without any slippage.
\end{itemize}
\item \textbf{Step 2: Placing the bread on the plate}
\begin{itemize}
\item \textbf{scoring 0:} The bread is not placed on the plate or is dropped.
\item \textbf{scoring 0.5:} The bread is placed on the plate but not accurately or takes multiple attempts.
\item \textbf{scoring 1:} The bread is accurately placed on the plate on the first attempt.
\end{itemize}
\item \textbf{Step 3: Grasping a slice of ham}
\begin{itemize}
\item \textbf{scoring 0:} The gripper does not grasp the ham.
\item \textbf{scoring 0.5:} The gripper attempts to grasp the ham multiple times, struggles to maintain a stable grasp, or takes too long to succeed.
\item \textbf{scoring 1:} The gripper successfully grasps the ham on the first attempt without any slippage.
\end{itemize}
\item \textbf{Step 4: Placing the ham on the bread}
\begin{itemize}
\item \textbf{scoring 0:} The ham is not placed on the bread or is dropped.
\item \textbf{scoring 0.5:} The ham is placed on the bread but not accurately or takes multiple attempts.
\item \textbf{scoring 1:} The ham is accurately placed on the bread on the first attempt.
\end{itemize}
\item \textbf{Step 5: Grasping a piece of lettuce}
\begin{itemize}
\item \textbf{scoring 0:} The gripper does not grasp the lettuce.
\item \textbf{scoring 0.5:} The gripper attempts to grasp the lettuce multiple times, struggles to maintain a stable grasp, or takes too long to succeed.
\item \textbf{scoring 1:} The gripper successfully grasps the lettuce on the first attempt without any slippage.
\end{itemize}
\item \textbf{Step 6: Placing the lettuce on the ham}
\begin{itemize}
\item \textbf{scoring 0:} The lettuce is not placed on the ham or is dropped.
\item \textbf{scoring 0.5:} The lettuce is placed on the ham but not accurately or takes multiple attempts.
\item \textbf{scoring 1:} The lettuce is accurately placed on the ham on the first attempt.
\end{itemize}
\item \textbf{Step 7: Grasping the second slice of bread}
\begin{itemize}
\item \textbf{scoring 0:} The gripper does not grasp the bread.
\item \textbf{scoring 0.5:} The gripper attempts to grasp the bread multiple times, struggles to maintain a stable grasp, or takes too long to succeed.
\item \textbf{scoring 1:} The gripper successfully grasps the bread on the first attempt without any slippage.
\end{itemize}
\item \textbf{Step 8: Placing the bread on the lettuce}
\begin{itemize}
\item \textbf{scoring 0:} The bread is not placed on the lettuce or is dropped.
\item \textbf{scoring 0.5:} The bread is placed on the lettuce but not accurately or takes multiple attempts.
\item \textbf{scoring 1:} The bread is accurately placed on the lettuce on the first attempt.
\end{itemize}
\end{itemize}

\subsubsection{Package Product}

\textbf{Task description.}
Package Product is a precision manipulation task that requires the robot to grasp a product and place it into a bag. The task consists of two steps: first, the robot must grasp the product; and second, the robot must place the product into the bag. The complexity of this task lies in the need for precise control over the grasping and placement actions, ensuring that the product is handled carefully and placed accurately into the bag.

\textbf{Scoring criteria.}
\begin{itemize}
\item \textbf{Step 1: Grasping the product}
\begin{itemize}
\item \textbf{scoring 0:} The gripper does not grasp the product.
\item \textbf{scoring 0.5:} The gripper attempts to grasp the product multiple times, struggles to maintain a stable grasp, or takes too long to succeed.
\item \textbf{scoring 1:} The gripper successfully grasps the product on the first attempt without any slippage.
\end{itemize}
\item \textbf{Step 2: Placing the product into the bag}
\begin{itemize}
\item \textbf{scoring 0:} The product is not placed into the bag or is dropped.
\item \textbf{scoring 0.5:} The product is placed into the bag but not accurately or takes multiple attempts.
\item \textbf{scoring 1:} The product is accurately placed into the bag on the first attempt.
\end{itemize}
\end{itemize}

\subsubsection{Clean Trash}

\textbf{Task description.}
Clean Trash is a precision manipulation task that requires the robot to grasp trash and place it into a trash bin. The task consists of two steps: first, the robot must grasp the trash; and second, the robot must place the trash into the trash bin. The complexity of this task lies in the need for precise control over the grasping and placement actions, ensuring that the trash is handled carefully and placed accurately into the trash bin.

\textbf{Scoring criteria.}
\begin{itemize}
\item \textbf{Step 1: Grasping the trash}
\begin{itemize}
\item \textbf{scoring 0:} The gripper does not grasp the trash.
\item \textbf{scoring 0.5:} The gripper attempts to grasp the trash multiple times, struggles to maintain a stable grasp, or takes too long to succeed.
\item \textbf{scoring 1:} The gripper successfully grasps the trash on the first attempt without any slippage.
\end{itemize}
\item \textbf{Step 2: Placing the trash into the trash bin}
\begin{itemize}
\item \textbf{scoring 0:} The trash is not placed into the trash bin or is dropped.
\item \textbf{scoring 0.5:} The trash is placed into the trash bin but not accurately or takes multiple attempts.
\item \textbf{scoring 1:} The trash is accurately placed into the trash bin on the first attempt.
\end{itemize}
\end{itemize}

\subsubsection{Industrial Sorting}

\textbf{Task description.}
Industrial Sorting is a precision manipulation task that requires the robot to identify two different items and place them into their corresponding designated areas using the left and right arms. The task consists of four steps: first, the right arm must grasp item 1; second, the right arm must place item 1 into its designated area; third, the left arm must grasp item 2; and fourth, the left arm must place item 2 into its designated area. The complexity of this task lies in the need for precise control over the grasping and placement actions, ensuring that each item is handled carefully and placed accurately into the correct area.

\textbf{Scoring criteria.}
\begin{itemize}
\item \textbf{Step 1: Right arm grasping item 1}
\begin{itemize}
\item \textbf{scoring 0:} The right gripper does not grasp item 1.
\item \textbf{scoring 0.5:} The right gripper attempts to grasp item 1 multiple times, struggles to maintain a stable grasp, or takes too long to succeed.
\item \textbf{scoring 1:} The right gripper successfully grasps item 1 on the first attempt without any slippage.
\end{itemize}
\item \textbf{Step 2: Right arm placing item 1 into its designated area}
\begin{itemize}
\item \textbf{scoring 0:} Item 1 is not placed into its designated area or is dropped.
\item \textbf{scoring 0.5:} Item 1 is placed into its designated area but not accurately or takes multiple attempts.
\item \textbf{scoring 1:} Item 1 is accurately placed into its designated area on the first attempt.
\end{itemize}
\item \textbf{Step 3: Left arm grasping item 2}
\begin{itemize}
\item \textbf{scoring 0:} The left gripper does not grasp item 2.
\item \textbf{scoring 0.5:} The left gripper attempts to grasp item 2 multiple times, struggles to maintain a stable grasp, or takes too long to succeed.
\item \textbf{scoring 1:} The left gripper successfully grasps item 2 on the first attempt without any slippage.
\end{itemize}
\item \textbf{Step 4: Left arm placing item 2 into its designated area}
\begin{itemize}
\item \textbf{scoring 0:} Item 2 is not placed into its designated area or is dropped.
\item \textbf{scoring 0.5:} Item 2 is placed into its designated area but not accurately or takes multiple attempts.
\item \textbf{scoring 1:} Item 2 is accurately placed into its designated area on the first attempt.
\end{itemize}
\end{itemize}

\subsubsection{\change{Push Chairs}}

\textbf{Task description.}
Push Chairs is a mobile manipulation task that requires the robot to coordinate base mobility with manipulation to tidy up a room by returning two chairs to their designated positions (e.g., under a table). The task consists of four steps: first, the robot must navigate to a position behind the first chair; second, the robot must push the first chair to its target location; third, the robot must navigate to a position behind the second chair; and fourth, the robot must push the second chair to its target location. The complexity of this task lies in the precise coupling of navigation and manipulation, requiring accurate alignment relative to the large objects to ensure effective pushing forces and correct final poses.

\textbf{Scoring criteria.}
\begin{itemize}
\item \textbf{Step 1: Moving behind the first chair}
\begin{itemize}
\item \textbf{scoring 0:} The robot fails to move or navigates to an incorrect location.
\item \textbf{scoring 0.5:} The robot moves towards the chair, but its orientation is significantly skewed or the distance is too large, rendering the subsequent pushing action impossible or highly unstable.
\item \textbf{scoring 1:} The robot successfully navigates to the correct position behind the first chair with proper alignment for pushing.
\end{itemize}
\item \textbf{Step 2: Pushing the first chair to the target position}
\begin{itemize}
\item \textbf{scoring 0:} The robot fails to push the chair, knocks it over, or pushes it to a completely incorrect area.
\item \textbf{scoring 0.5:} The robot pushes the chair, but the final orientation is significantly misaligned or the chair remains far from the specific target position.
\item \textbf{scoring 1:} The robot successfully pushes the first chair to the designated position with correct orientation on the first attempt.
\end{itemize}
\item \textbf{Step 3: Moving behind the second chair}
\begin{itemize}
\item \textbf{scoring 0:} The robot fails to move or navigates to an incorrect location.
\item \textbf{scoring 0.5:} The robot moves towards the chair, but its orientation is significantly skewed or the distance is too large, rendering the subsequent pushing action impossible or highly unstable.
\item \textbf{scoring 1:} The robot successfully navigates to the correct position behind the second chair with proper alignment for pushing.
\end{itemize}
\item \textbf{Step 4: Pushing the second chair to the target position}
\begin{itemize}
\item \textbf{scoring 0:} The robot fails to push the chair, knocks it over, or pushes it to a completely incorrect area.
\item \textbf{scoring 0.5:} The robot pushes the chair, but the final orientation is significantly misaligned or the chair remains far from the specific target position.
\item \textbf{scoring 1:} The robot successfully pushes the second chair to the designated position with correct orientation on the first attempt.
\end{itemize}
\end{itemize}

%% file: root.bbl
\begin{thebibliography}{10}
\providecommand{\url}[1]{#1}
\csname url@samestyle\endcsname
\providecommand{\newblock}{\relax}
\providecommand{\bibinfo}[2]{#2}
\providecommand{\BIBentrySTDinterwordspacing}{\spaceskip=0pt\relax}
\providecommand{\BIBentryALTinterwordstretchfactor}{4}
\providecommand{\BIBentryALTinterwordspacing}{\spaceskip=\fontdimen2\font plus
\BIBentryALTinterwordstretchfactor\fontdimen3\font minus \fontdimen4\font\relax}
\providecommand{\BIBforeignlanguage}[2]{{%
\expandafter\ifx\csname l@#1\endcsname\relax
\typeout{** WARNING: IEEEtran.bst: No hyphenation pattern has been}%
\typeout{** loaded for the language `#1'. Using the pattern for}%
\typeout{** the default language instead.}%
\else
\language=\csname l@#1\endcsname
\fi
#2}}
\providecommand{\BIBdecl}{\relax}
\BIBdecl

\bibitem{go1}
Q.~Bu, J.~Cai, L.~Chen, X.~Cui, Y.~Ding, S.~Feng, S.~Gao, X.~He, X.~Huang, S.~Jiang \emph{et~al.}, ``{AgiBot World Colosseo}: A large-scale manipulation platform for scalable and intelligent embodied systems,'' \emph{arXiv preprint arXiv:2503.06669}, 2025.

\bibitem{gpt4}
OpenAI, ``{GPT}-4 {T}echnical {R}eport,'' \emph{arXiv preprint arXiv:2303.08774}, 2023.

\bibitem{gemini}
R.~Anil, S.~Borgeaud, J.-B. Alayrac, J.~Yu, R.~Soricut, J.~Schalkwyk, A.~M. Dai, A.~Hauth, K.~Millican \emph{et~al.}, ``Gemini: a family of highly capable multimodal models,'' \emph{arXiv preprint arXiv:2312.11805}, 2023.

\bibitem{sam2}
N.~Ravi, V.~Gabeur, Y.-T. Hu, R.~Hu, C.~Ryali, T.~Ma, H.~Khedr, R.~R{\"a}dle, C.~Rolland, L.~Gustafson \emph{et~al.}, ``{SAM} 2: Segment anything in images and videos,'' \emph{arXiv preprint arXiv:2408.00714}, 2024.

\bibitem{clip}
A.~Radford, J.~W. Kim, C.~Hallacy, A.~Ramesh, G.~Goh, S.~Agarwal, G.~Sastry, A.~Askell, P.~Mishkin, J.~Clark \emph{et~al.}, ``Learning transferable visual models from natural language supervision,'' in \emph{ICML}, 2021.

\bibitem{llava}
H.~Liu, C.~Li, Q.~Wu, and Y.~J. Lee, ``Visual instruction tuning,'' in \emph{NeurIPS}, 2023.

\bibitem{blip2}
J.~Li, D.~Li, S.~Savarese, and S.~Hoi, ``{BLIP}-2: Bootstrapping language-image pre-training with frozen image encoders and large language models,'' in \emph{ICML}, 2023.

\bibitem{internvl}
Z.~Chen, J.~Wu, W.~Wang, W.~Su, G.~Chen, S.~Xing, M.~Zhong, Q.~Zhang, X.~Zhu, L.~Lu \emph{et~al.}, ``{InternVL}: Scaling up vision foundation models and aligning for generic visual-linguistic tasks,'' in \emph{CVPR}, 2024.

\bibitem{qwen25vl}
S.~Bai, K.~Chen, X.~Liu, J.~Wang, W.~Ge, S.~Song, K.~Dang, P.~Wang, S.~Wang, J.~Tang \emph{et~al.}, ``{Qwen2.5-VL} technical report,'' \emph{arXiv preprint arXiv:2502.13923}, 2025.

\bibitem{prismatic}
S.~Karamcheti, S.~Nair, A.~Balakrishna, P.~Liang, T.~Kollar, and D.~Sadigh, ``{Prismatic VLMs}: Investigating the design space of visually-conditioned language models,'' in \emph{ICML}, 2024.

\bibitem{vit}
A.~Dosovitskiy, L.~Beyer, A.~Kolesnikov, D.~Weissenborn, X.~Zhai, T.~Unterthiner, M.~Dehghani, M.~Minderer, G.~Heigold, S.~Gelly \emph{et~al.}, ``An image is worth 16x16 words: Transformers for image recognition at scale,'' in \emph{ICLR}, 2021.

\bibitem{mae}
K.~He, X.~Chen, S.~Xie, Y.~Li, P.~Doll{\'a}r, and R.~Girshick, ``Masked autoencoders are scalable vision learners,'' in \emph{CVPR}, 2022.

\bibitem{dinov2}
M.~Oquab, T.~Darcet, T.~Moutakanni, H.~Vo, M.~Szafraniec, V.~Khalidov, P.~Fernandez, D.~Haziza, F.~Massa, A.~El-Nouby \emph{et~al.}, ``{DINO}v2: Learning robust visual features without supervision,'' \emph{TMLR}, 2024.

\bibitem{brohan2023rt2}
A.~Brohan, N.~Brown, J.~Carbajal, Y.~Chebotar, X.~Chen, K.~Choromanski, T.~Ding, D.~Driess, A.~Dubey, C.~Finn \emph{et~al.}, ``{RT-2}: Vision-language-action models transfer web knowledge to robotic control,'' in \emph{CoRL}, 2023.

\bibitem{kim2024openvla}
M.~J. Kim, K.~Pertsch, S.~Karamcheti, T.~Xiao, A.~Balakrishna, S.~Nair, R.~Rafailov, E.~Foster, G.~Lam, P.~Sanketi \emph{et~al.}, ``{OpenVLA}: An open-source vision-language-action model,'' in \emph{CoRL}, 2024.

\bibitem{pi0}
K.~Black, N.~Brown, D.~Driess, A.~Esmail, M.~Equi, C.~Finn, N.~Fusai, L.~Groom, K.~Hausman, B.~Ichter \emph{et~al.}, ``{\(\pi\)}\({}_{\mbox{0}}\): {A} vision-language-action flow model for general robot control,'' in \emph{RSS}, 2025.

\bibitem{rdt}
S.~Liu, L.~Wu, B.~Li, H.~Tan, H.~Chen, Z.~Wang, K.~Xu, H.~Su, and J.~Zhu, ``{RDT-1B}: a diffusion foundation model for bimanual manipulation,'' in \emph{ICLR}, 2025.

\bibitem{univla}
Q.~Bu, Y.~Yang, J.~Cai, S.~Gao, G.~Ren, M.~Yao, P.~Luo, and H.~Li, ``{UniVLA}: Learning to act anywhere with task-centric latent actions,'' in \emph{RSS}, 2025.

\bibitem{gr00t}
J.~Bjorck, F.~Casta{\~n}eda, N.~Cherniadev, X.~Da, R.~Ding, L.~Fan, Y.~Fang, D.~Fox, F.~Hu, S.~Huang \emph{et~al.}, ``{GR00T} {N1:} an open foundation model for generalist humanoid robots,'' \emph{arXiv preprint arXiv:2503.14734}, 2025.

\bibitem{chen2025position}
L.~Chen, C.~Sima, K.~Chitta, A.~Loquercio, P.~Luo, Y.~Ma, and H.~Li, ``{Intelligent Robot Manipulation Requires Self-Directed Learning},'' \emph{Authorea Preprints}, 2025.

\bibitem{bridgev1}
F.~Ebert, Y.~Yang, K.~Schmeckpeper, B.~Bucher, G.~Georgakis, K.~Daniilidis, C.~Finn, and S.~Levine, ``{Bridge Data}: Boosting generalization of robotic skills with cross-domain datasets,'' in \emph{RSS}, 2022.

\bibitem{walke2023bridgedata}
H.~R. Walke, K.~Black, T.~Z. Zhao, Q.~Vuong, C.~Zheng, P.~Hansen-Estruch, A.~W. He, V.~Myers, M.~J. Kim, M.~Du \emph{et~al.}, ``{BridgeData} v2: A dataset for robot learning at scale,'' in \emph{CoRL}, 2023.

\bibitem{droid}
A.~Khazatsky, K.~Pertsch, S.~Nair, A.~Balakrishna, S.~Dasari, S.~Karamcheti, S.~Nasiriany, M.~K. Srirama, L.~Y. Chen, K.~Ellis \emph{et~al.}, ``{DROID}: A large-scale in-the-wild robot manipulation dataset,'' in \emph{RSS}, 2024.

\bibitem{oxe}
A.~Padalkar, A.~Pooley, A.~Jain, A.~Bewley, A.~Herzog, A.~Irpan, A.~Khazatsky, A.~Rai, A.~Singh, A.~Brohan \emph{et~al.}, ``{Open X-Embodiment}: Robotic learning datasets and {RT-X} models,'' in \emph{ICRA}, 2024.

\bibitem{gaoyanglaw}
F.~Lin, Y.~Hu, P.~Sheng, C.~Wen, J.~You, and Y.~Gao, ``Data scaling laws in imitation learning for robotic manipulation,'' in \emph{ICLR}, 2025.

\bibitem{manibox}
H.~Tan, X.~Xu, C.~Ying, X.~Mao, S.~Liu, X.~Zhang, H.~Su, and J.~Zhu, ``{ManiBox}: Enhancing spatial grasping generalization via scalable simulation data generation,'' \emph{arXiv preprint arXiv:2411.01850}, 2024.

\bibitem{wang2024scaling}
L.~Wang, X.~Chen, J.~Zhao, and K.~He, ``Scaling proprioceptive-visual learning with heterogeneous pre-trained transformers,'' in \emph{NeurIPS}, 2024.

\bibitem{zheng2025universal}
J.~Zheng, J.~Li, D.~Liu, Y.~Zheng, Z.~Wang, Z.~Ou, Y.~Liu, J.~Liu, Y.-Q. Zhang, and X.~Zhan, ``Universal actions for enhanced embodied foundation models,'' in \emph{CVPR}, 2025.

\bibitem{yang2024pushing}
J.~Yang, C.~Glossop, A.~Bhorkar, D.~Shah, Q.~Vuong, C.~Finn, D.~Sadigh, and S.~Levine, ``Pushing the limits of cross-embodiment learning for manipulation and navigation,'' in \emph{RSS}, 2024.

\bibitem{doshi2024scaling}
R.~Doshi, H.~Walke, O.~Mees, S.~Dasari, and S.~Levine, ``{Scaling Cross-Embodied Learning}: One policy for manipulation, navigation, locomotion and aviation,'' in \emph{CoRL}, 2024.

\bibitem{li2025train}
H.~Li, Y.~Cui, and D.~Sadigh, ``How to train your robots? the impact of demonstration modality on imitation learning,'' in \emph{ICRA}, 2025.

\bibitem{chi2023diffusion}
C.~Chi, S.~Feng, Y.~Du, Z.~Xu, E.~Cousineau, B.~Burchfiel, and S.~Song, ``{Diffusion Policy}: Visuomotor policy learning via action diffusion,'' in \emph{RSS}, 2023.

\bibitem{bert}
J.~Devlin, M.-W. Chang, K.~Lee, and K.~Toutanova, ``{BERT}: Pre-training of deep bidirectional transformers for language understanding,'' in \emph{NAACL}, 2019.

\bibitem{llama}
H.~Touvron, T.~Lavril, G.~Izacard, X.~Martinet, M.-A. Lachaux, T.~Lacroix, B.~Rozi{\`e}re, N.~Goyal, E.~Hambro, F.~Azhar \emph{et~al.}, ``{LLaMA}: Open and efficient foundation language models,'' \emph{arXiv preprint arXiv:2302.13971}, 2023.

\bibitem{gpt3}
T.~Brown, B.~Mann, N.~Ryder, M.~Subbiah, J.~D. Kaplan, P.~Dhariwal, A.~Neelakantan, P.~Shyam, G.~Sastry, A.~Askell \emph{et~al.}, ``Language models are few-shot learners,'' in \emph{NeurIPS}, 2020.

\bibitem{minigpt4}
D.~Zhu, J.~Chen, X.~Shen, X.~Li, and M.~Elhoseiny, ``{MiniGPT-4}: Enhancing vision-language understanding with advanced large language models,'' in \emph{ICLR}, 2024.

\bibitem{li2023roboflamingo}
X.~Li, M.~Liu, H.~Zhang, C.~Yu, J.~Xu, H.~Wu, C.~Cheang, Y.~Jing, W.~Zhang, H.~Liu, H.~Li, and T.~Kong, ``Vision-language foundation models as effective robot imitators,'' in \emph{ICLR}, 2024.

\bibitem{driess2023palm}
D.~Driess, F.~Xia, M.~S. Sajjadi, C.~Lynch, A.~Chowdhery, B.~Ichter, A.~Wahid, J.~Tompson, Q.~Vuong, T.~Yu \emph{et~al.}, ``{PaLM-E}: An embodied multimodal language model,'' in \emph{ICML}, 2023.

\bibitem{wen2025tinyvla}
J.~Wen, Y.~Zhu, J.~Li, M.~Zhu, Z.~Tang, K.~Wu, Z.~Xu, N.~Liu, R.~Cheng, C.~Shen \emph{et~al.}, ``{TinyVLA}: Towards fast, data-efficient vision-language-action models for robotic manipulation,'' \emph{RA-L}, 2025.

\bibitem{wen2025dexvla}
J.~Wen, Y.~Zhu, J.~Li, Z.~Tang, C.~Shen, and F.~Feng, ``{DexVLA}: Vision-language model with plug-in diffusion expert for general robot control,'' in \emph{CoRL}, 2025.

\bibitem{brohan2022rt1}
A.~Brohan, N.~Brown, J.~Carbajal, Y.~Chebotar, J.~Dabis, C.~Finn, K.~Gopalakrishnan, K.~Hausman, A.~Herzog, J.~Hsu \emph{et~al.}, ``{RT-1}: Robotics transformer for real-world control at scale,'' in \emph{RSS}, 2023.

\bibitem{team2024octo}
D.~Ghosh, H.~Walke, K.~Pertsch, K.~Black, O.~Mees, S.~Dasari, J.~Hejna, T.~Kreiman, C.~Xu \emph{et~al.}, ``{Octo}: An open-source generalist robot policy,'' in \emph{RSS}, 2024.

\bibitem{something2something}
R.~Goyal, S.~Ebrahimi~Kahou, V.~Michalski, J.~Materzynska, S.~Westphal, H.~Kim, V.~Haenel, I.~Fruend, P.~Yianilos, M.~Mueller-Freitag \emph{et~al.}, ``The" something something" video database for learning and evaluating visual common sense,'' in \emph{ICCV}, 2017.

\bibitem{ego4d}
K.~Grauman, A.~Westbury, E.~Byrne, Z.~Chavis, A.~Furnari, R.~Girdhar, J.~Hamburger, H.~Jiang, M.~Liu, X.~Liu \emph{et~al.}, ``{Ego4D}: Around the world in 3,000 hours of egocentric video,'' in \emph{CVPR}, 2022.

\bibitem{genie}
J.~Bruce, M.~D. Dennis, A.~Edwards, J.~Parker-Holder, Y.~Shi, E.~Hughes, M.~Lai, A.~Mavalankar, R.~Steigerwald, C.~Apps \emph{et~al.}, ``Genie: Generative interactive environments,'' in \emph{ICML}, 2024.

\bibitem{lapa}
S.~Ye, J.~Jang, B.~Jeon, S.~Joo, J.~Yang, B.~Peng, A.~Mandlekar, R.~Tan, Y.-W. Chao, B.~Y. Lin \emph{et~al.}, ``Latent action pretraining from videos,'' in \emph{ICLR}, 2025.

\bibitem{igor}
X.~Chen, J.~Guo, T.~He, C.~Zhang, P.~Zhang, D.~C. Yang, L.~Zhao, and J.~Bian, ``{IGOR}: Image-goal representations are the atomic control units for foundation models in embodied ai,'' \emph{arXiv preprint arXiv:2411.00785}, 2024.

\bibitem{gr1}
H.~Wu, Y.~Jing, C.~Cheang, G.~Chen, J.~Xu, X.~Li, M.~Liu, H.~Li, and T.~Kong, ``Unleashing large-scale video generative pre-training for visual robot manipulation,'' in \emph{ICLR}, 2024.

\bibitem{gr2}
C.-L. Cheang, G.~Chen, Y.~Jing, T.~Kong, H.~Li, Y.~Li, Y.~Liu, H.~Wu, J.~Xu, Y.~Yang \emph{et~al.}, ``{GR-2}: A generative video-language-action model with web-scale knowledge for robot manipulation,'' \emph{arXiv preprint arXiv:2410.06158}, 2024.

\bibitem{unipi}
Y.~Du, S.~Yang, B.~Dai, H.~Dai, O.~Nachum, J.~Tenenbaum, D.~Schuurmans, and P.~Abbeel, ``Learning universal policies via text-guided video generation,'' in \emph{NeurIPS}, 2023.

\bibitem{mpi}
J.~Zeng, Q.~Bu, B.~Wang, W.~Xia, L.~Chen, H.~Dong, H.~Song, D.~Wang, D.~Hu, P.~Luo \emph{et~al.}, ``Learning manipulation by predicting interaction,'' in \emph{RSS}, 2024.

\bibitem{clover}
Q.~Bu, J.~Zeng, L.~Chen, Y.~Yang, G.~Zhou, J.~Yan, P.~Luo, H.~Cui, Y.~Ma, and H.~Li, ``Closed-loop visuomotor control with generative expectation for robotic manipulation,'' in \emph{NeurIPS}, 2024.

\bibitem{dit}
W.~Peebles and S.~Xie, ``Scalable diffusion models with transformers,'' in \emph{ICCV}, 2023.

\bibitem{rh20t}
H.-S. Fang, H.~Fang, Z.~Tang, J.~Liu, C.~Wang, J.~Wang, H.~Zhu, and C.~Lu, ``{RH20T}: A comprehensive robotic dataset for learning diverse skills in one-shot,'' in \emph{ICRA}, 2024.

\bibitem{ario}
Z.~Wang, H.~Zheng, Y.~Nie, W.~Xu, Q.~Wang, H.~Ye, Z.~Li, K.~Zhang, X.~Cheng, W.~Dong \emph{et~al.}, ``{All Robots in One}: A new standard and unified dataset for versatile, general-purpose embodied agents,'' \emph{arXiv preprint arXiv:2408.10899}, 2024.

\bibitem{robomind}
K.~Wu, C.~Hou, J.~Liu, Z.~Che, X.~Ju, Z.~Yang, M.~Li, Y.~Zhao, Z.~Xu, G.~Yang \emph{et~al.}, ``{RoboMIND}: Benchmark on multi-embodiment intelligence normative data for robot manipulation,'' in \emph{RSS}, 2025.

\bibitem{wu2025freetacman}
L.~Wu, C.~Yu, J.~Ren, L.~Chen, R.~Huang, G.~Gu, and H.~Li, ``{FreeTacMan}: Robot-free visuo-tactile data collection system for contact-rich manipulation,'' in \emph{ICRA}, 2026.

\bibitem{pan2025ams}
Y.~Pan, R.~Qiao, L.~Chen, K.~Chitta, L.~Pan, H.~Mai, Q.~Bu, C.~Zheng, H.~Zhao, P.~Luo, and H.~Li, ``{Agility Meets Stability}: Versatile humanoid control with heterogeneous data,'' in \emph{ICRA}, 2026.

\bibitem{team2025gemini}
G.~R. Team, A.~Abdolmaleki, S.~Abeyruwan, J.~Ainslie, J.-B. Alayrac, M.~G. Arenas, A.~Balakrishna, N.~Batchelor, A.~Bewley, J.~Bingham \emph{et~al.}, ``{Gemini Robotics 1.5}: Pushing the frontier of generalist robots with advanced embodied reasoning, thinking, and motion transfer,'' \emph{arXiv preprint arXiv:2510.03342}, 2025.

\bibitem{kareer2025emergence}
S.~Kareer, K.~Pertsch, J.~Darpinian, J.~Hoffman, D.~Xu, S.~Levine, C.~Finn, and S.~Nair, ``Emergence of human to robot transfer in vision-language-action models,'' \emph{arXiv preprint arXiv:2512.22414}, 2025.

\bibitem{saxena2025matters}
V.~Saxena, M.~Bronars, N.~R. Arachchige, K.~Wang, W.~C. Shin, S.~Nasiriany, A.~Mandlekar, and D.~Xu, ``What matters in learning from large-scale datasets for robot manipulation,'' in \emph{ICLR}, 2025.

\bibitem{hejna2408re}
J.~Hejna, C.~A. Bhateja, Y.~Jiang, K.~Pertsch, and D.~Sadigh, ``{ReMix}: Optimizing data mixtures for large scale imitation learning,'' in \emph{CoRL}, 2024.

\bibitem{zheng2024preliminary}
Y.~Zheng, Z.~Xia, Q.~Zhang, T.~Zhang, B.~Lu, X.~Huo, C.~Han, Y.~Li, M.~Yu, B.~Jin \emph{et~al.}, ``Preliminary investigation into data scaling laws for imitation learning-based end-to-end autonomous driving,'' \emph{arXiv preprint arXiv:2412.02689}, 2024.

\bibitem{maniskill}
T.~Mu, Z.~Ling, F.~Xiang, D.~Yang, X.~Li, S.~Tao, Z.~Huang, Z.~Jia, and H.~Su, ``{ManiSkill}: Generalizable manipulation skill benchmark with large-scale demonstrations,'' in \emph{NeurIPS Datasets and Benchmarks}, 2021.

\bibitem{robotwin}
Y.~Mu, T.~Chen, S.~Peng, Z.~Chen, Z.~Gao, Y.~Zou, L.~Lin, Z.~Xie, and P.~Luo, ``Robotwin: Dual-arm robot benchmark with generative digital twins (early version),'' in \emph{ECCV}, 2025.

\bibitem{masuya2025variable}
N.~Masuya, S.~Sakaino, and T.~Tsuji, ``Variable-frequency imitation learning for variable-speed motion,'' in \emph{ICM}, 2025.

\bibitem{zhai2023siglip}
X.~Zhai, B.~Mustafa, A.~Kolesnikov, and L.~Beyer, ``Sigmoid loss for language image pre-training,'' in \emph{ICCV}, 2023.

\bibitem{guo2025demospeedup}
L.~Guo, Z.~Xue, Z.~Xu, and H.~Xu, ``{DemoSpeedup}: Accelerating visuomotor policies via entropy-guided demonstration acceleration,'' in \emph{CoRL}, 2025.

\bibitem{arachchige2025sail}
N.~R. Arachchige, Z.~Chen, W.~Jung, W.~C. Shin, R.~Bansal, P.~Barroso, Y.~H. He, Y.~C. Lin, B.~Joffe, S.~Kousik \emph{et~al.}, ``{SAIL}: Faster-than-demonstration execution of imitation learning policies,'' in \emph{CoRL}, 2025.

\end{thebibliography}
